%% file: feats.tex
\documentclass[letterpaper, 10 pt, conference]{ieeeconf}  %
\IEEEoverridecommandlockouts                              %
\input{includes}
\usepackage{comment}
\newcommand\scalemath[2]{\scalebox{#1}{\mbox{\ensuremath{\displaystyle #2}}}}
\newcommand{\firstrank}[1]{\textcolor{blue}{#1}}
\newcommand{\secondrank}[1]{\textcolor{orange}{#1}}
\newcommand{\thirdrank}[1]{\textcolor{violet}{#1}}

\let\ACMmaketitle=\maketitle
\renewcommand{\maketitle}{\begingroup\let\footnote=\thanks \ACMmaketitle\endgroup}

\makeatletter
\newcommand*\titleheader[1]{\begingroup\gdef\@titleheader{#1}\let\footnote=\thanks\endgroup}
\AtBeginDocument{%
  \let\st@red@title\@title
  \def\@title{%
  \begin{flushleft}
    \vspace{-2.15em}
    \bgroup\normalfont\small\@titleheader\par\egroup
    \vspace{-15pt}\par\noindent\rule{\textwidth}{0.1pt}
    \end{flushleft}
    \vskip0.0em\st@red@title
        }

}

\usepackage{fancyhdr}
\newcommand{\mytitle}{
\copyright 2025 IEEE. Personal use of this material is permitted. Permission from IEEE must be obtained for all other uses, including reprinting/republishing this material for advertising or promotional purposes, collecting new collected works for resale or redistribution to servers or lists, or reuse of any copyrighted component of this work in other works.}

\makeatother

\title{\LARGE \bf
Learning Force Distribution Estimation for the GelSight Mini\\ Optical Tactile Sensor Based on Finite Element Analysis
}

\titleheader{\textbf{Accepted version.} Published in the 2025 IEEE/RSJ International Conference on Intelligent Robots and Systems (IROS 2025). \\ DOI: 10.1109/IROS60139.2025.11246486}

\author{Erik Helmut$^{*}\-^{1}$, Luca Dziarski$^{*}\-^{2}$, Niklas Funk$^{2}$, Boris Belousov$^{3}$, Jan Peters$^{2,3,4,5}$%
\thanks{*Authors contributed equally.}%
\thanks{Corresponding author: Erik Helmut. Email: erik.helmut1@gmail.com.}%
\thanks{$^{1}$Department of Computational Engineering, Technical University of Darmstadt \quad $^{2}$Department of Computer Science, Technical University of Darmstadt \quad $^{3}$German Research Center for AI (DFKI) \quad $^{4}$ Centre for Cognitive Science, Technical University of Darmstadt \quad $^{5}$ Hessian Center for Artificial Intelligence (Hessian.AI), Darmstadt}%
\thanks{
This work received funding from ``The Adaptive Mind'' grant, the EU’s Horizon Europe project ARISE (Grant no.: 101135959), the AICO grant by the Nexplore/Hochtief Collaboration with TU Darmstadt, and from the Hessisches Ministerium für Wissenschaft \& Kunst through the DFKI grant.}%
}

\begin{document}
\input{acronyms}
\maketitle
\thispagestyle{fancy}

\begin{abstract}
Contact-rich manipulation remains a major challenge in robotics. Optical tactile sensors like GelSight Mini offer a low-cost solution for contact sensing by capturing soft-body deformations of the silicone gel. However, accurately inferring shear and normal force distributions from these gel deformations has yet to be fully addressed. In this work, we propose a machine learning approach using a U-net architecture to predict force distributions directly from the sensor's raw images. Our model, trained on force distributions inferred from \ac{fea}, demonstrates promising accuracy in predicting normal and shear force distributions for the commercially available GelSight Mini sensor.
It also shows potential for generalization across indenters, sensors of the same type, and for enabling real-time application. 
The codebase, dataset and models are open-sourced and available at \url{https://feats-ai.github.io}.
 \end{abstract}

\section{Introduction}
\begin{figure}[t]
    \centering
    \includegraphics[width=0.45\textwidth]{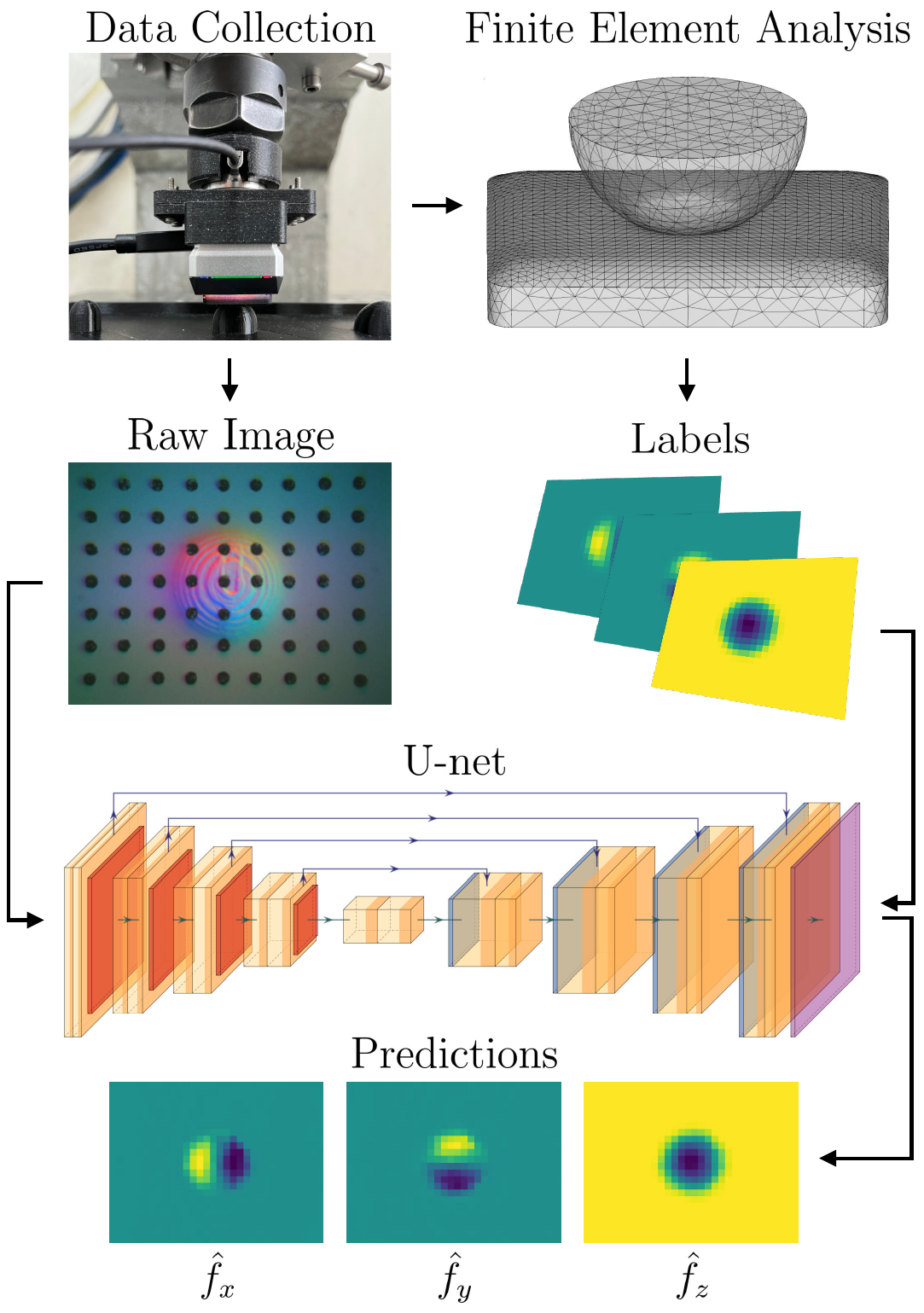}
    \caption{Method Overview from data collection to force distribution prediction. Starting from collecting versatile indentation data in a precisely calibrated setup with a CNC milling machine, we employ Finite Element Analysis for label generation, i.e., obtaining the corresponding ``ground truth'' force distributions. Through the combination of labels and the raw tactile images, we train a U-net capable of efficiently mapping raw tactile images to the corresponding force distributions during inference.}
    \label{fig:summary}
\end{figure}

Tactile sensing plays an important role in advancing the state-of-the-art in robotic manipulation \cite{harmon1982automated, harmon1984tactile, billard2019trends, abad_visuotactile_2020, lach2023placing}.
Successful applications include grip adaptation through slip detection \cite{kaboli_tactile-based_2016, sui_incipient_2021, yussof_sensorization_2010, funk2024evetac}, medical procedures \cite{tiwana_review_2012, othman_tactile_2022} and tele-operation \cite{zhu_visual_2022}. 
In particular, optical tactile sensors have emerged as a promising technology for capturing contact information due to their high spatial resolution, multimodal sensing capabilities~\cite{higuera2024sparsh}—including shape \cite{ito2010shape}, hardness \cite{yuan2017shape}, texture \cite{fang2018dual}, and temperature \cite{sun2019novel}—and cost-effectiveness \cite{dahiya_tactile_2010, othman_tactile_2022}. However, many prior works have focused on extracting only low-dimensional tactile information, such as total force \cite{9981100, Li2021Ftouch, Naeini2020Dynamic}, limiting operational flexibility. Access to contact force distributions, on the other hand, would enable better handling of multiple contacts and diverse manipulation scenarios \cite{sferrazza2020simlearning}.

Conventional methods for extracting force distributions require calculating the three-dimensional deformation of the contact medium and utilizing elasticity theory \cite{kamiyama2004evaluation, ma_dense_2019, van2020large, zhang_tac3d_2022}. Yet, accounting for non-linear material behavior, such as with \acf{fea}, is computationally intensive and unsuitable for real-time applications.

Recent works leverage Deep Learning to address the challenge of real-time force estimation.
In~\cite{yuan_gelsight_2017}, Convolutional Neural Networks (CNNs) were used to predict contact forces from sensor images, while \cite{funk_canfnet_2023} introduced CANFnet for estimating normal force distributions at the pixel level.
In~\cite{sferrazza2020simlearning, sferrazza-unet, sferrazza_ground_2019, xue2023acquisition}, \ac{fea}-derived data was used to train a model for predicting force distributions, demonstrating the effectiveness of combining simulations with data-driven methods.
Yet,~\cite{sferrazza_ground_2019,sferrazza2020simlearning, sferrazza-unet,xue2023acquisition} rely on custom-made optical tactile sensors, thereby imposing a significantly increased entry barrier into the field as the models are not compatible with off-the-shelf, commercially available sensors.

Inspired by these works, this paper introduces FEATS (see Fig.~\ref{fig:summary})---a machine learning approach directly mapping raw tactile images to force distributions.
FEATS is tailored to the commercially available GelSight Mini optical tactile sensor~\cite{yuan_gelsight_2017, gelSightMinionline}, thereby dropping the need for sensor manufacturing and drastically extending the applicability of the approach.
Furthermore, this sensor allows for a significantly expanded range of measurable forces $\unit[0 - 40]{N}$, an $8$-fold increase in the maximum measurable force compared to~\cite{sferrazza_ground_2019}, and an $10$-fold increase compared to~\cite{xue2023acquisition}.
FEATS is trained through an extensive dataset of real-world indentation data and the corresponding force labels obtained from the open-source FEA solver CalculiX~\cite{ccxDocumentation}.
To ensure high data quality, we collect the data using a CNC milling machine.
On the representation level, FEATS leverages an efficient U-net neural network architecture \cite{ronneberger_u-net_2015}, which is trained to map from a single GelSight image to the corresponding force distribution during inference.
We optimize the network architecture for providing accurate force predictions while maintaining low inference times suitable for real-time control.
Our thorough experimental results demonstrate that FEATS accurately predicts high-dimensional contact force distributions from raw tactile images and generalizes to novel, previously unseen indenters.
Moreover, we showcase that the underlying U-Net architecture outperforms several baselines varying w.r.t. network size and output resolution.

To summarize, our key contributions are i) a method for creating force labels from FEA outputs tailored to the commercially available GelSight Mini, ii) the collection of a large dataset of versatile indentation data, iii) a model architecture capable of generalizing across objects and allowing for real-time inference, and iv) open-sourcing our dataset, the code for data collection, label generation, model training, and the trained models for aiding reuse and reproducibility.

\section{Related Work}
Extracting meaningful contact-related information from the raw RGB images of optical tactile sensors is a major challenge in visual-tactile perception \cite{zhu_visual_2022, kamiyama2004evaluation, funk_canfnet_2023, sferrazza_ground_2019, li_marker_2023, johnson2009retrographic}.
A number of methods have been proposed for constructing or learning such ``tactile representations''.

\subsection{Marker Displacement Methods}
\label{sec:marker-displacement-methods}
Li et al. \cite{li_marker_2023} posit that it is the contact layer deformations that capture the crucial information within tactile images. By analyzing these deformations, various contact features can be extracted, with \acp{mdm} \cite{li_marker_2023}. 
\Acp{mdm} rely on markers within the elastomer which appear as features in the sensor's image (Fig. \ref{fig:summary}).
For the GelSight sensor~\cite{yuan_gelsight_2017, gelSightMinionline}, markers were first introduced in \cite{yuan_measurement_2015} to study normal and shear forces, along with slip dynamics.
They identified a linear relationship between loads and marker motion, but this applied only in non-slip conditions.
Beyond marker motion, optical sensors can capture detailed height maps and contact geometry through careful illumination and photometric stereo \cite{johnson2009retrographic}.
These height maps can be used to estimate contact forces with a third-degree polynomial~\cite{zhu_visual_2022}.
In this paper, we use a gel with markers, but their movement is not explicitly tracked.
Instead, they serve as implicit features within the sensor image, which is analyzed by a neural network to predict force distributions.

\subsection{Learning-Based Tactile Representations}
\label{sec:optical-tactile-sensors-and-deep-learning}
Advancements in computer vision directly translate to vision-based tactile sensing.
Models such as CNNs, LSTMs, and SVMs were adapted to assess object hardness~\cite{yuan2017shape}, grip stability~\cite{li2018slip}, and lump detection~\cite{jia2013lump}.
More tactile-specific deep learning methods have been developed for overall force~\cite{yuan_gelsight_2017} and pixel-wise contact area and normal force estimation~\cite{funk_canfnet_2023}.

Building on the demonstrated effectiveness of deep neural networks for feature extraction and prediction, we employ a U-net architecture similar to that of~\cite{funk_canfnet_2023}.
However, in contrast to~\cite{funk_canfnet_2023}, FEATS estimates both normal and shear forces, thus providing a physically grounded representation in the form of a 3D force distribution acting upon the sensor.

\subsection{Force Distribution Prediction Through Elasticity Theory}
\begin{figure*}[t]
    \vspace{2mm}
    \centering
    \includegraphics[width=0.9\textwidth, trim={0cm 1.2cm 0 1cm}, clip]{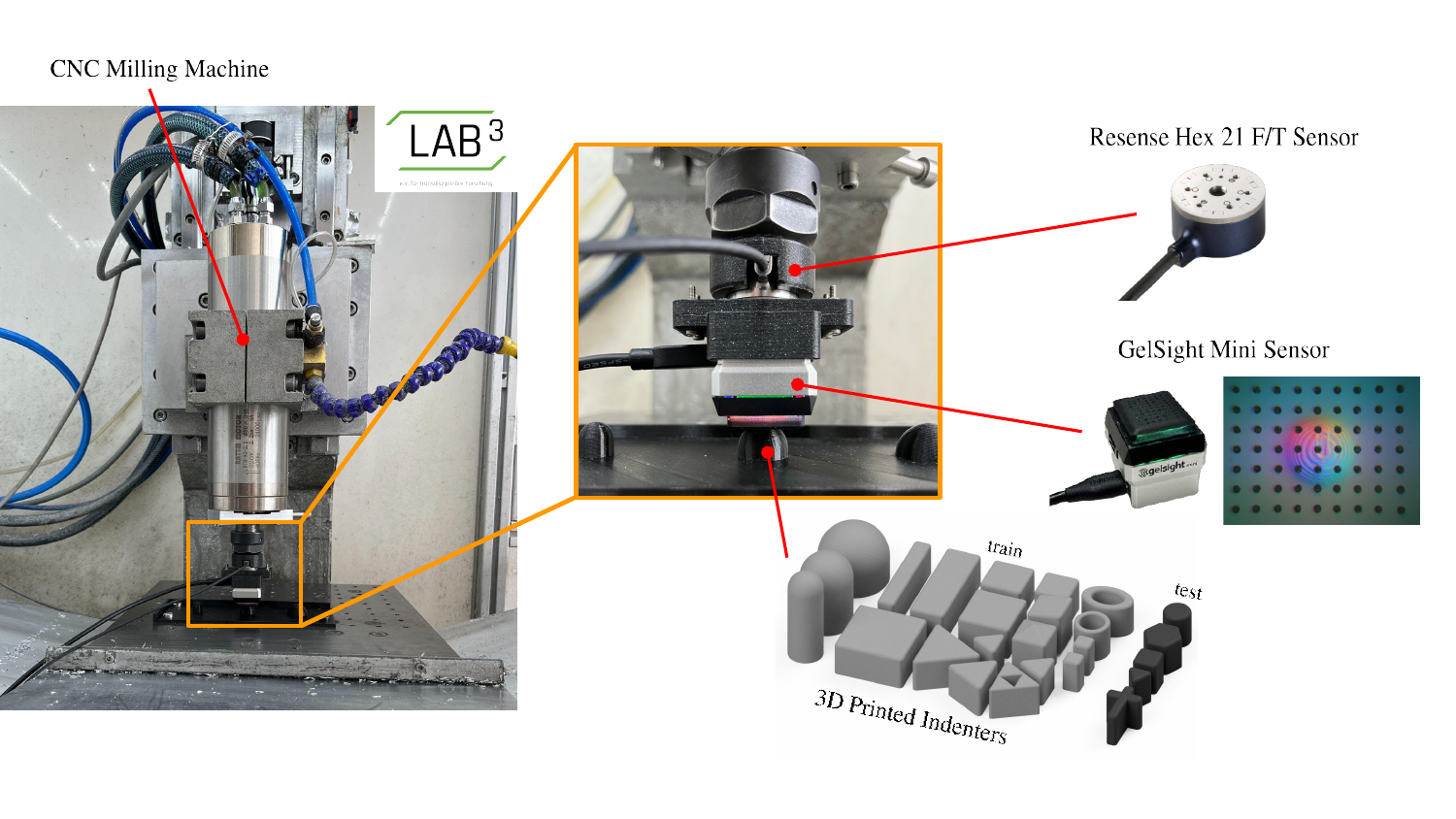}
    \caption{Experimental setup for data collection. We use a CNC milling machine to create different calibrated contact configurations between the GelSight Mini tactile sensor and 3D printed indenters. We also place a six-axis F/T sensor above the tactile sensor to obtain complementary external force measurements.
    }
    \label{fig:indenters}
\end{figure*}

Elasticity theory has been effectively applied to create more refined and accurate load distributions acting upon the soft silicone gel of optical tactile sensors.
In~\cite{kamiyama2004evaluation}, elasticity theory with \acp{mdm} was used to derive force vectors from marker movements assuming a linear elastic, uniform and half-spaced material.
This method was later adopted in~\cite{ma_dense_2019} for the GelSlim sensor~\cite{donlon2018gelslim}.
More recently, sensors enabling 3D surface deformation reconstruction have been proposed, such as TacLINK~\cite{van2020large} and Tac3D~\cite{zhang_tac3d_2022}. 
They compute force distributions from measured 3D marker displacements.

However, direct prediction of force distributions from displacements, usually through a linear stiffness matrix, does not account for the nonlinearities of soft elastomers. Sun et al. \cite{sun_soft_2022} addressed this limitation by training a ResNet \cite{he_deep_2015} on sensor images with approximated force distributions using a thumb-sized vision-based tactile sensor called Insight. Andrussow et al.~\cite{andrussov_minisight} presented a miniaturized successor, Minisight, which retains the same force labeling methodology in a fingertip-sized form factor. In contrast, Sferrazza et al. \cite{sferrazza_ground_2019} utilized a neural network trained on image features with force distributions obtained from \ac{fea}, providing more physically accurate labels. They subsequently improved this approach using simulated training data \cite{sferrazza2020simlearning, sferrazza-unet}.
The use of FEA-generated labels was further demonstrated by Xue et al.~\cite{xue2023acquisition}, which also considered interactions with soft objects. This methodology was later adopted for larger tactile surfaces, with Luu et al.~\cite{luu_large_vitac} applying it to the TacLINK~\cite{van2020large} sensor, and Ho and Nakayama~\cite{Ho03062021} incorporating it in their IoTouch system.

Building on these approaches, we also aim to estimate contact forces using supervised deep learning.
However, in contrast to the aforementioned approaches relying on pipelines designed for non-public sensors and involving preprocessing (e.g., optical flow~\cite{sferrazza-unet}) or sim-to-real transfer~\cite{luu_large_vitac}, we use raw RGB images from the commercially available GelSight Mini sensor.
Crucially, we develop specific procedures for data collection and model training that enable the efficient use of this widely accessible sensor, thereby significantly lowering the entry barrier into the field.

Our key contributions are: i) method for creating force labels from FEA outputs tailored to GelSight Mini + implementation in CalculiX, ii) data collection procedure + dataset, iii) trained model applicable to varying objects and different sensors of the same type.

\section{Method}
Our proposed method---Finite Element Analysis for Tactile Sensing (FEATS)---estimates the force distribution acting upon the gel of the GelSight Mini sensor by approximating the output of a Finite Element Analysis (FEA) computation with a neural network that takes the raw RGB image as input.
Querying a neural network is much faster than running FEA, and importantly, for FEA computation, one needs a precise geometrical description of the contact, whereas FEATS only needs a raw input image and no further geometric information.
Thus, FEATS exhibits low inference times and does not require any additional object tracking equipment at run time.
Only at training time, FEATS requires a labeled dataset collected in a controlled environment on a set of calibration objects, so-called indenters (cf. Fig.~\ref{fig:indenters}).

\subsection{Data Collection}
\label{sec:data_collection}
The data collection process involves a series of precise indentation experiments.
We attach a GelSight Mini sensor to the fixed spindle of a \ac{cnc} milling machine, with a positional tolerance of $\pm \unit[0.25]{\mu m}$ to automatically create random contact configurations between the sensor and selected indenters (cf. Fig.~\ref{fig:indenters}).
This work leverages $24$ 3-d printed indenters ($19$ for training and $5$ for testing) varying in shape and size as shown at the bottom of Fig.~\ref{fig:indenters}.
The optical tactile sensor, i.e., Gelsight Mini, is equipped with the dotted gel to better track the indentation motion.
We additionally place a six-axis \textit{RESENSE-HEX-21} Force/Torque (F/T) sensor above the Gelsight Mini to have a complementary external force measurement, which is later used to validate the material model.
With this setup, we collected a total of $14279$ samples. Each sample contains the respective GelSight Mini RGB image, the \ac{cnc} motion data, i.e., the current indentation, and the F/T sensor's readings.
The forces in the $z$ direction reach up to $\unit[40]{N}$, and in the $x$ and $y$ directions up to $\pm \unit[5]{N}$.

\begin{figure*}[t]
\vspace{1mm}
    \centering
    \includegraphics[width=0.9\textwidth, trim={0cm 5cm 0 2cm}, clip]{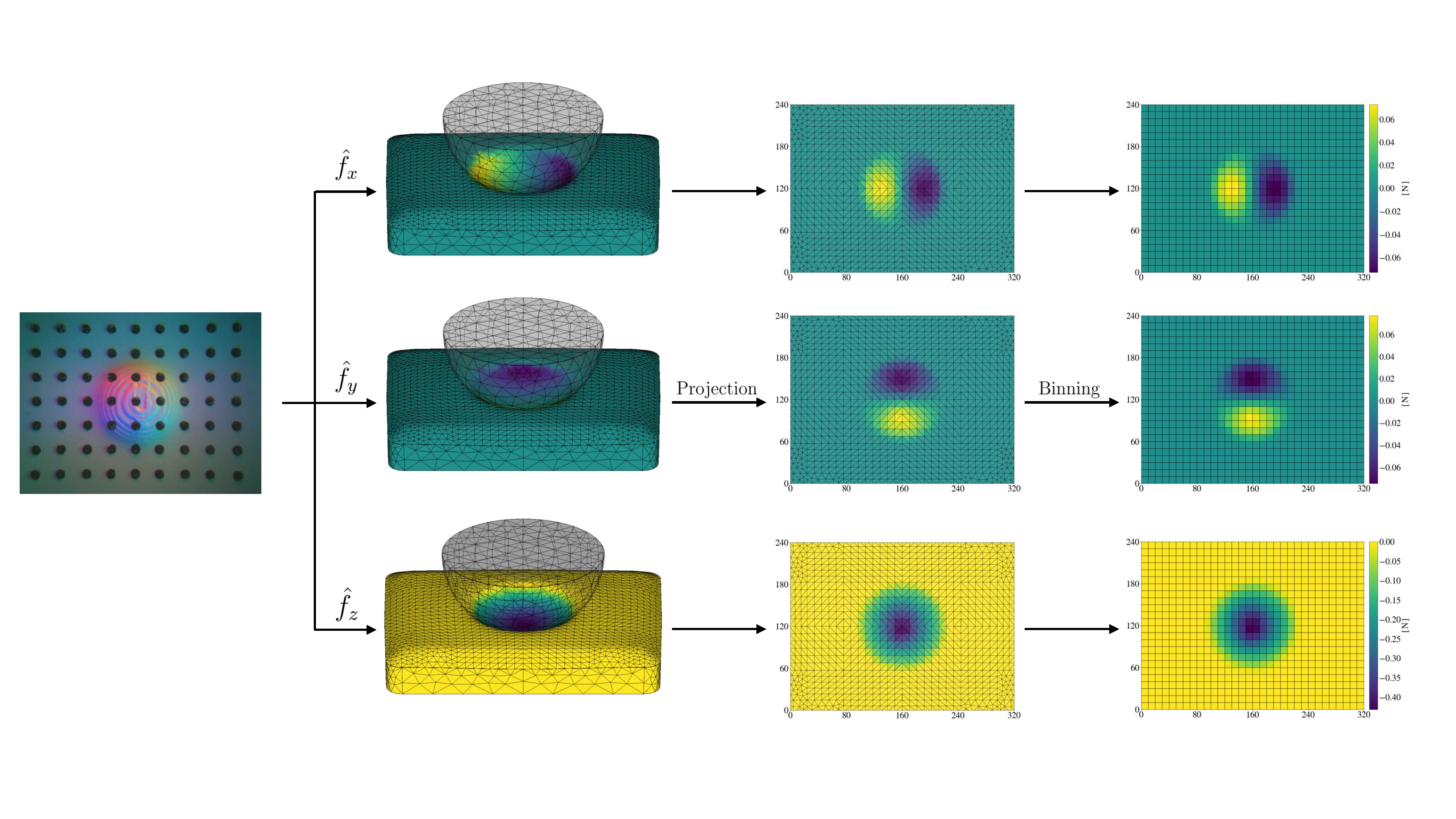}
    \caption{Label creation process when a spherical indenter presses into the sensor's soft silicone gel. From left to right: 1) GelSight Mini image, 2) Simulation of the contact configuration and the raw output from running the 3D FEA, 3) Projecting the result from the 3D FEA into the coordinate system of Gelsight Mini, i.e., into an image plane, 4) Force labels after changing the resolution to $24 \times 32$.
    }
    \label{fig:creating_labels}
\end{figure*}

\subsection{Finite Element Analysis}
\label{sec:finite_element_analysis}
For simulating the indentation experiments, we employ \acf{fea} using the open-source solver CalculiX \cite{ccxDocumentation}. The solver's capability for nonlinear computations makes it an appropriate tool for the stress analysis of soft elastomers. Consequently, it can calculate the resulting force distributions corresponding to the previously described real-world indentation experiments -- an essential component for label generation as described in the next Section \ref{sec:creating_labels}.
We generate tetrahedral volume meshes for both the gel and indenters, with denser meshing at the contact surface to ensure higher accuracy of the simulated contact forces (see Fig.~\ref{fig:creating_labels}). The gel mesh comprises $2504$ contact surface elements out of a total of $7591$ elements.
The elements are implemented as ten-node tetrahedral (C3D10 \cite{ccxDocumentation}) elements.
The FEA is conducted as a static analysis, assuming hard contact and applying tied contact constraints.
This approach permits the execution of simulations without requiring a friction coefficient, and is substantiated by the fact that the experiments are conducted in a manner that precluded slippage.
Assuming no deformation of the indenters, they are characterized as a hard material with a Young's modulus of \unit[210]GPa and a Poisson ratio of 0.3.
In \cite{yuan_gelsight_2017}, the gel elastomer of the GelSight sensor is characterized as a material similar to a Neo-Hookean solid with a shear modulus, $\mu$, of $0.145$.
When employing a hyperelastic Neo-Hookean model in the CalculiX solver \cite{ccxDocumentation}, the strain energy potential is expressed as $U = C_{10} (\bar{I}_1 - 3) + 1/D_1 (J - 1)^{2}$ where $\bar{I}_1$ denotes the first invariant of the right Cauchy-Green deformation tensor, while $J$ represents the determinant of the deformation gradient tensor. $C_{10}$ and $D_1$ are the material constants to be set.
In agreement with \cite{yuan_gelsight_2017}, $C_{10}$ is chosen to be $0.0725$.
Due to CalculiX's lack of support for perfectly incompressible materials, the solver assigns a default value to $D_1$ prior to the simulation.
This model is later validated in the results section (cf. Sec. \ref{sec:material_characterization}) by optimizing the material parameters through load-depth indentation data \cite{chen_macroindentation_2013}, comparing the normal forces from the \ac{fea} with F/T sensor measurements.

\subsection{Creating Labels}
\label{sec:creating_labels}
To obtain the approximated ``ground truth'' force distribution labels, each real-world indentation experiment (cf. Sec. \ref{sec:data_collection}) is repeated in simulation in order to calculate the contact forces using \ac{fea} (cf. Sec. \ref{sec:finite_element_analysis} \& Fig.~\ref{fig:creating_labels}).
For each element on the surface of the gel, three force components are computed: shear forces in the $x$ and $y$ directions, and normal forces in the $z$ direction.

\paragraph{Mesh Projection} To align the results from the three-dimensional \ac{fea} with the sensor's raw images, the 3D coordinates of the gel mesh nodes are initially projected onto the 2D image plane of the GelSight Mini through $\mathbf{x}^{\text{GS}} = \mathbf{P} \cdot \mathbf{X}^{\text{FEA}}$, where $\mathbf{X}^{\text{FEA}}$ are the 3D coordinates of the mesh nodes, $\mathbf{x}^{\text{GS}}$ the image projections and $\mathbf{P}$ the projection matrix. The projection matrix $\mathbf{P}$ can be determined using the least squares method. Four point correspondences are needed for a minimal solution.
To establish point correspondences, we gently press the GelSight Mini against an object, such as a cuboid of known size, using the CNC milling machine.
Once the object's shape is clearly visible in the raw sensor image, we identify distinctive object points, such as its corners, in the image.
We subsequently repeat the same indentation experiment in simulation and also extract the corner points in the coordinates of the simulation.
This leaves us with the point correspondences needed for calculating $\mathbf{P}$.

\paragraph{Discretization Process} Once all nodes have been projected onto the image plane, the force distributions for shear and normal forces are binned within the image boundaries. The number of bins decides the resolution of the force distribution. In most experiments, it is $24 \times 32$, however, the resolution can be seamlessly adapted.
Each bin’s force value is calculated by summing the contact force contributions of all elements intersecting that bin (see Fig. \ref{fig:creating_labels}), where the element position refers to its state before deformation. The contribution of each element to a bin is proportional to the fraction of the element that falls within the bin.
This process is repeated for each force component.
Note that all the code for label generation is open-sourced on our website.

\subsection{U-Net for Learning Force Distribution Estimation}
\label{sec:unet_force_distribution_estimation}
\begin{figure}[t]
    \centering
    \vspace{3mm}
    \includegraphics[width=\columnwidth]{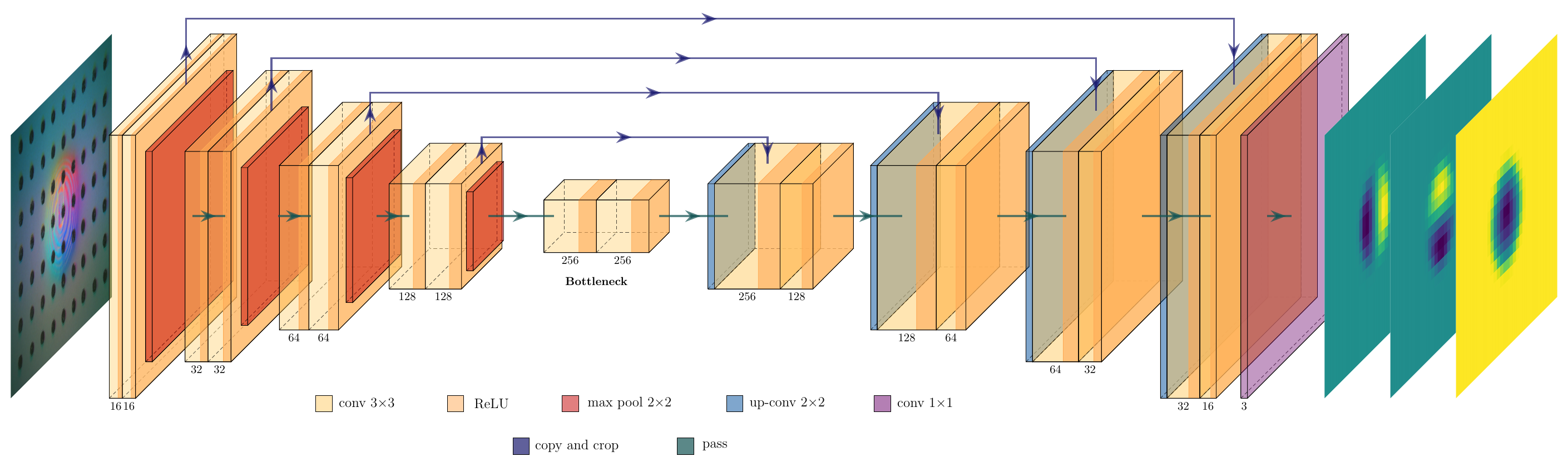}
    \caption[]{U-net model which maps raw images from GelSight Mini sensor to shear and normal force distributions. The architecture comprises $4$ down-sampling (encoder) and $4$ up-sampling (decoder) blocks, connected by skip connections. The number of feature channels at each stage is labeled at the bottom of the corresponding block. Different colors of the boxes and arrows indicate specific operations and activation functions. This image is generated using PlotNeuralNet~\cite{haris_iqbal_2018_2526396}.}
    \label{fig:unet}
\end{figure}

To predict shear and normal force distributions from the raw GelSight images, we employ a U-net architecture~\cite{ronneberger_u-net_2015}, which is well-suited for spatially-detailed tasks.
The network's encoder-decoder structure allows for efficiently mapping raw sensor images to force distributions, as demonstrated by \cite{sferrazza-unet}.
The model takes $240 \times 320$ RGB images as input and outputs three force maps for both shear force components and the normal forces (see Fig. \ref{fig:unet}).

\paragraph{Architecture}
The U-net follows an encoder-decoder structure, consisting of a contracting path (encoder) and an expansive path (decoder).
High-level image features are extracted by the encoder, by iteratively applying two $3 \times 3$ convolutions with zero padding, followed by a ReLU activation function, and a $2 \times 2$ max-pooling with a stride of $2$.
This leads to down-sampling while progressively doubling the feature channels at each step. The decoder up-samples the feature map at each step and then applies a $2 \times 2$ convolution, which halves the number of feature channels but doubles the resolution. The up-sampled feature map is then concatenated with the corresponding cropped feature map from the encoder. Subsequently, two $3 \times 3$ convolutions are applied, both of which are followed by a ReLU activation.
In the final decoder layer, the last feature map is transformed into the specified number of classes via $1 \times 1$ convolution.

\paragraph{Training}
The model is trained by minimizing the \ac{mse} loss
\begin{equation}
\scalemath{0.9}{
\label{eq:loss}
    \text{MSE}(\mathbf{f}, \mathbf{\hat{f}}) = \frac{1}{3\text{W}\text{H}} \sum_{i=0}^{\text{W}} \sum_{j=0}^{\text{H}} \| \mathbf{f}^{(i,j)} - \mathbf{\hat{f}}^{(i,j)} \|_2^2
    }
\end{equation}
between predicted ($\mathbf{\hat{f}}$) and ground truth ($\mathbf{{f}}$) \ac{fea} force distribution components.
The \ac{mse} is averaged across all entries of the force distribution grid.

We train the model using $9084$ samples and use a validation dataset containing $538$ samples. We later evaluate the model on three different test datasets for which we provide the details in the respective experimental sections.
Prior to training, both input images and the ground truth force distribution labels are normalized with respect to the training dataset to ensure a consistent baseline across the data. Shear forces are normalized to a range of $-1$ to $+1$, while normal forces range from $0$ to $+1$. Data augmentation techniques, such as adding Gaussian noise and adjusting image brightness, contrast, saturation, and hue, are used to enhance the model's generalization capabilities. Training is carried out using the Adam optimizer \cite{kingma_adam_2017}, with an initial learning rate of $0.001$ and a batch size of $8$. The learning rate is adjusted adaptively based on validation performance. As validation loss, the \ac{mae} on the sum of total forces in the unnormalized space is used.
We train for $400$ epochs and select the model with the lowest validation loss.

\section{Results}
This section evaluates the performance of our proposed FEATS method.
The employed FEA model, which is the basis for our label generation procedure, has one free parameter $C_{10}$ that characterizes the response of the material to shear stress and needs to be set correctly.
Therefore, we start by validating a value for $C_{10}$ previously reported for GelSight~\cite{yuan_gelsight_2017} for applicability to GelSight Mini
(Sec.~\ref{sec:material_characterization}).
Subsequently, we evaluate the U-net accuracy in predicting shear and normal force distributions, along with its ability to reconstruct total forces (Sec.~\ref{sec:force_distribution_estimation}).
Particular attention is given to how different force distribution resolutions impact total force reconstruction accuracy.
We end with assessing the U-net's inference speed (Sec.~\ref{sec:inference_speed}).

\subsection{Material Characterization}
\label{sec:material_characterization}
To validate the Neo-Hookean material model parameter~$C_{10}$, load-depth indentation data can be utilized (cf. Sec. \ref{sec:finite_element_analysis}). The load-depth measurements are obtained, using a sphere indenter with a $\unit[15]{mm}$ diameter, by sampling data points at different indentation depths ranging from \unit[0.5]{mm} to \unit[2.0]{mm}, increasing in \unit[0.5]{mm} intervals.
The measured normal forces from the F/T sensor in our experimental setup $\tilde{f}^{(i)}_z$ are compared with the forces $f^{(i)}_z(C_{10})$ calculated by the \ac{fea} via the MSE loss \linebreak \( J(C_{10}) = 1/N \sum_{i=1}^{N} \left(f^{(i)}_{z}(C_{10}) - \tilde{f}^{(i)}_{z} \right)^2 \).
When running Bayesian optimization to find the best fit for these $N=4$ measurements, we found that $\hat{C}_{10} = 0.0792$ provided the best fit with a \ac{mae} of $\unit[0.5166]{N}$ between \ac{fea} and F/T sensor measurements.
Thus, we confirm that our estimate is within the range of $C_{10} = 0.0725$ previously reported in~\cite{yuan_gelsight_2017} for the GelSight sensor.
In the experiments, we use that value as it was stated in the original GelSight paper and our estimate closely matches it.

\begin{figure}[t]
    \centering
    \rotatebox[origin=l]{90}{\makebox[1.1in]{$x$ component}}%
    \hskip 0.5em
    \subfloat{\includegraphics[height=7.7em]{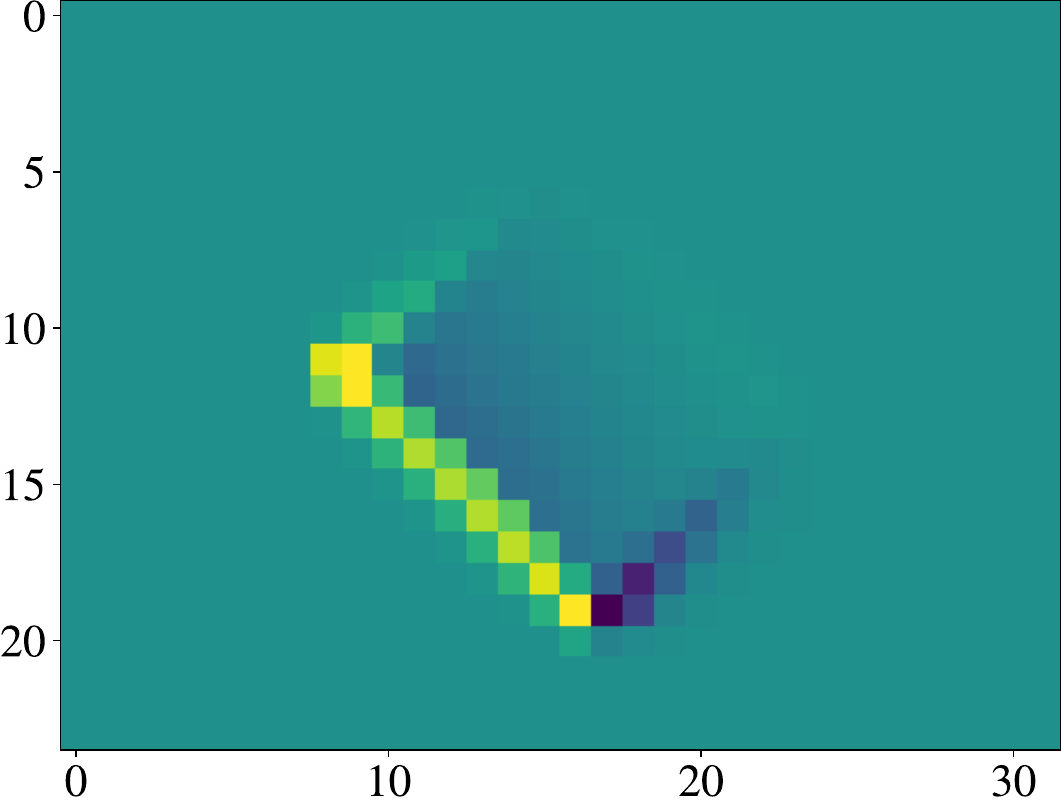}} \hskip 0.5em
    \subfloat{\includegraphics[height=7.7em]{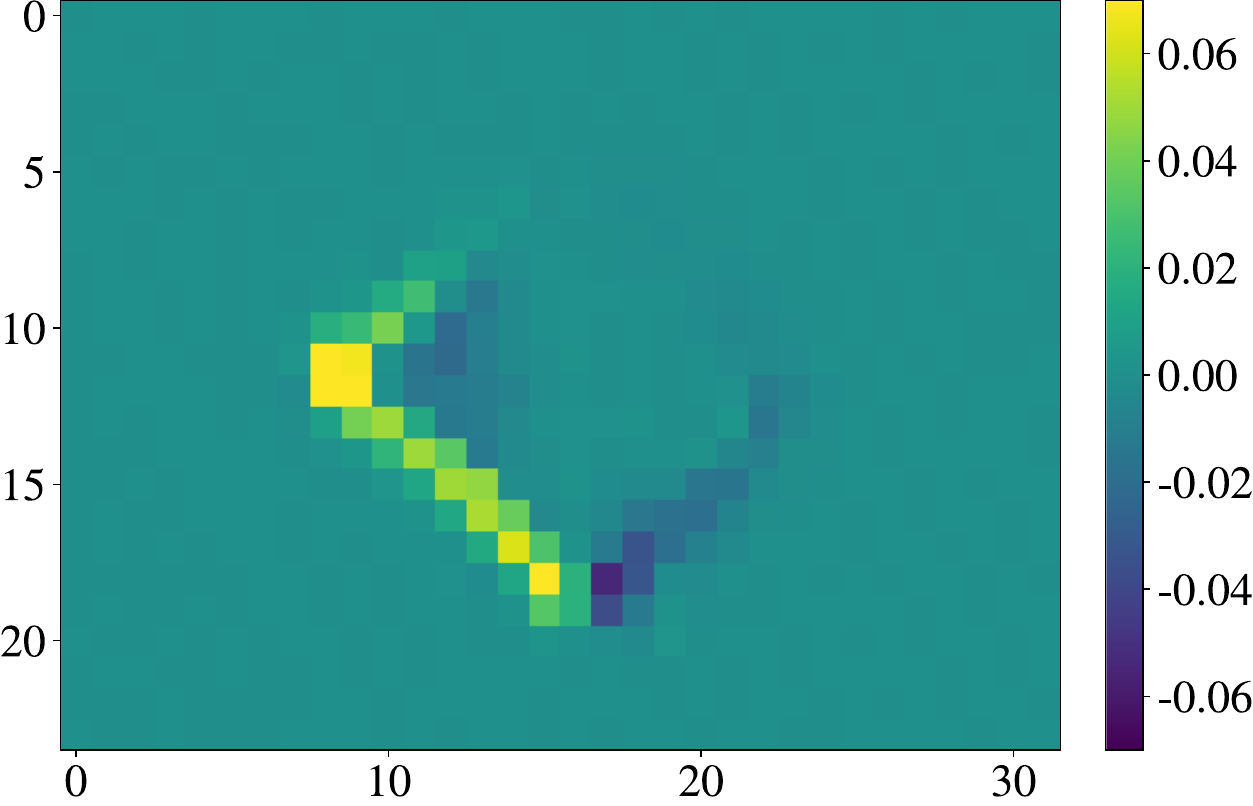}} \vspace{-0.5em}

    \rotatebox[origin=l]{90}{\makebox[1.1in]{$y$ component}}%
    \hskip 0.5em
    \subfloat{\includegraphics[height=7.7em]{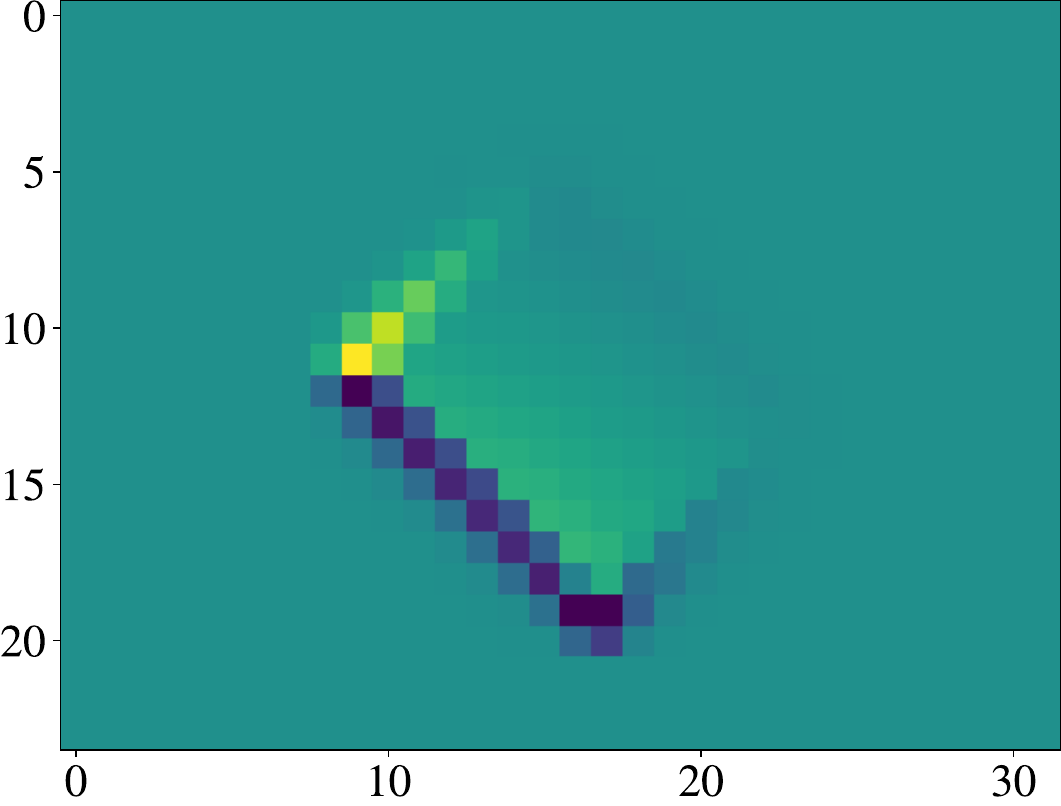}} \hskip 0.5em
    \subfloat{\includegraphics[height=7.7em]{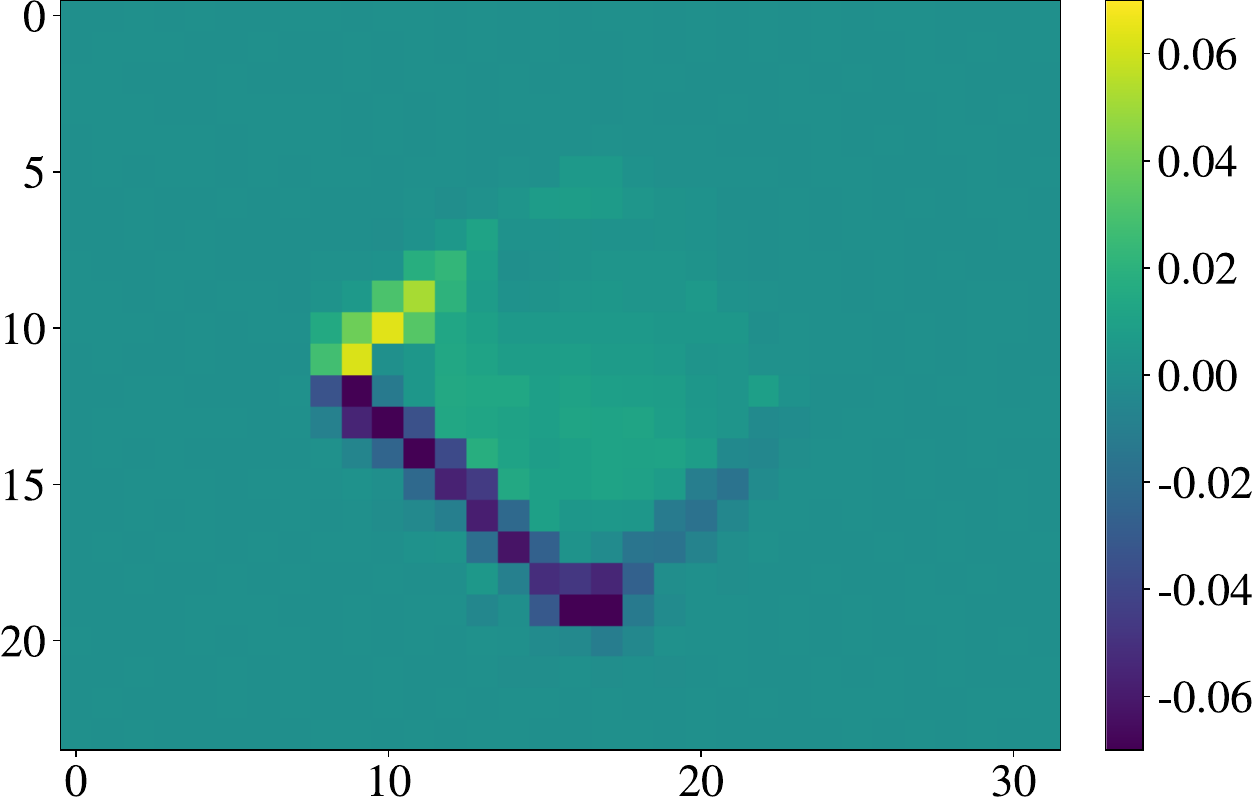}} \vspace{-0.5em}

    \rotatebox[origin=l]{90}{\makebox[1.1in]{$z$ component}}%
    \hskip 0.5em
    \subfloat{\includegraphics[height=7.7em]{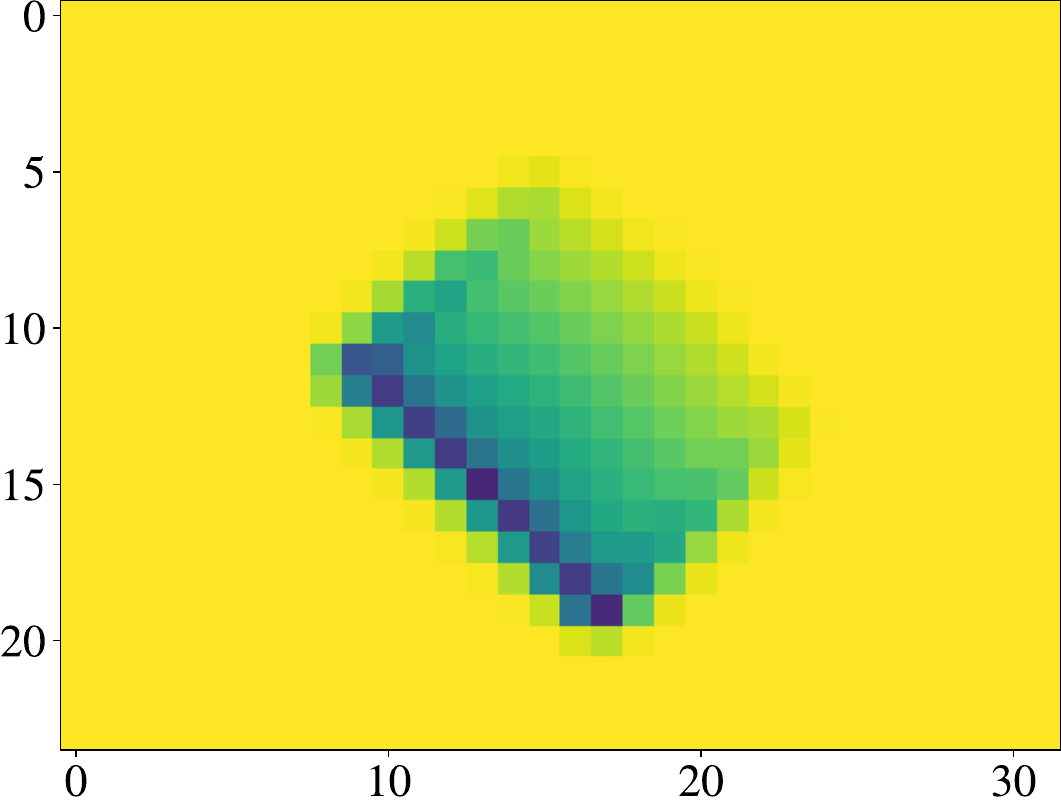}} \hskip 0.5em
    \subfloat{\includegraphics[height=7.7em]{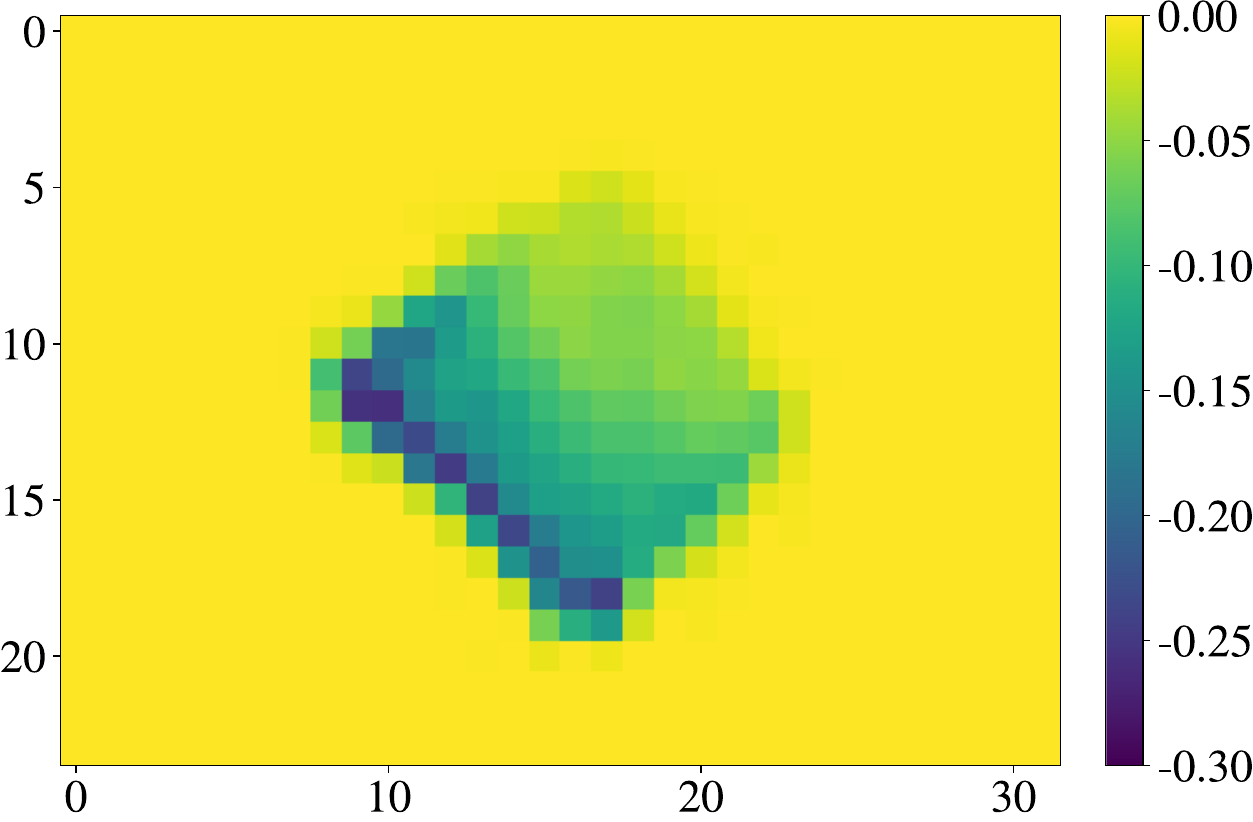}} \vspace{-0.2em}

    \makebox[1.3in]{ground truth} \hskip 0.5em \makebox[1.3in]{prediction}
    
    \caption{
    Ground truth labels (left column) and predictions of our FEATS model (right column) of the force distributions, in Newtons.
    The sloping cuboid indenter (cf. Fig.~\ref{fig:indenters}) penetrates the gel for $\unit[1.23]{mm}$, exerting a significant normal force.
    The resulting gel deformation causes shear forces, in this case roughly cancelling each other due to the absence of horizontal movement of the indenter.
    }
    \label{fig:pred_label_grid}
\end{figure}

\subsection{Evaluation of U-net Force Distribution Estimation}
\label{sec:force_distribution_estimation}

Next, we assess the U-nets' capacity to accurately estimate shear and normal force distributions from the raw RGB images of a GelSight Mini sensor.
We evaluate the trained networks on multiple test datasets that cover a wide range of forces and contact configurations.
The test dataset collection follows the procedure described in Secs.~\ref{sec:data_collection} \& \ref{sec:creating_labels}).

The following experimental evaluation is split into four parts and aims to answer the following questions: (a) How well does the method perform on a test dataset containing different indentations with the same indenters as seen during training?, (b) How does the network architecture and output resolution affect the force prediction accuracy?, (c) Can the network generalize to novel, previously unseen testing indenters?, and (d) Can the network generalize to a different GelSight Mini sensor?

Throughout all the experiments, we quantitatively evaluate the model performance through reporting the \ac{mae} between the predicted and ground truth force values, both for per-Grid Unit Forces (GUF) and for the Total Force (TF) values, i.e.,

\begin{equation}
\scalemath{0.9}{
    \text{MAE}_{\text{GUF}}(\mathbf{f}, \mathbf{\hat{f}}) = \frac{1}{3\text{WH}}\sum_{i=0}^{\text{W}} \sum_{j=0}^{\text{H}} \left|\mathbf{f}^{(i,j)} - \hat{\mathbf{f}}^{(i,j)} \right|
}
\end{equation}

\begin{equation}
\scalemath{0.9}{
    \text{MAE}_{\text{TF}}(\mathbf{f}, \mathbf{\hat{f}}) = \left| \sum_{i=0}^{\text{W}} \sum_{j=0}^{\text{H}} \mathbf{f}^{(i,j)} - \sum_{i=0}^{\text{W}} \sum_{j=0}^{\text{H}} \hat{\mathbf{f}}^{(i,j)} \right| \: .
    }
\end{equation}

We are reporting the GUF errors in addition to the TF errors since they directly reflect the error in the models' individual output values.
They, therefore, also capture spatial information regarding the contact area, whereas TF is missing out on the local information.

\paragraph{Evaluation of U-Net Predictions on the Test Dataset}

\begin{table}[b]
\vspace{-1em}
\centering
\caption{U-net Mean Absolute Error (MAE) on the Test Set containing known Indenters.}
\label{tab:evaluation_unet}
\begin{tabular}{lcc}
\multicolumn{1}{c|}{} & $\text{MAE}_{\text{GUF}}$ {[}N{]} & $\text{MAE}_{\text{TF}}$ {[}N{]} \\ \hline
\multicolumn{1}{l|}{$f_x$} & $0.0010 \pm 0.0013$ & $0.2932 \pm 0.4173$ \\
\multicolumn{1}{l|}{$f_y$} & $0.0010 \pm 0.0012$ & $0.2302 \pm 0.3405$ \\
\multicolumn{1}{l|}{$f_z$} & $0.0030 \pm 0.0029$ & $0.7994 \pm 1.0144$ \\ \hline
\end{tabular}
\end{table}

\begin{figure}
    \vspace{1em}
    \centering
    \includegraphics[width=.95\linewidth, trim={0 0 4.8cm 0}, clip]{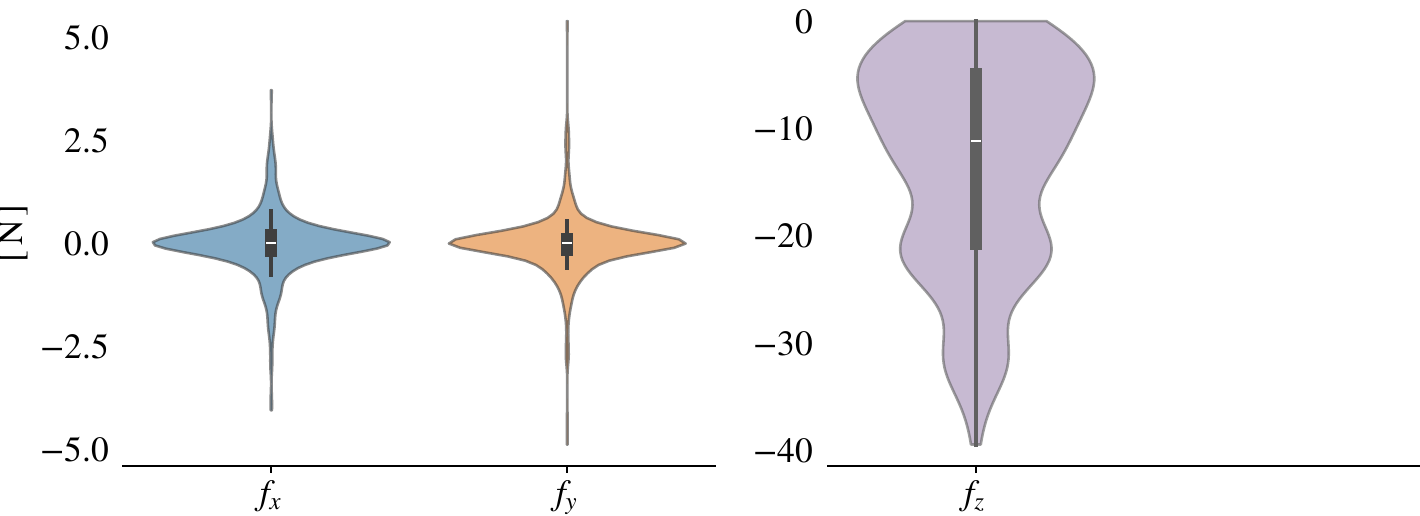}
    \caption{Test data ranges for the Total Force (TF) estimation task.
    The range of the tested normal forces is considerably larger than for the shear forces.
    }
    \label{fig:enter-label}
\end{figure}

This first evaluation considers our main U-net model that is trained to output the force distribution with dimensions $24 \times 32 \times 3$, therefore striking a good balance between maintaining a sufficiently high spatial resolution and at the same time low error on the test data.
The test dataset contains $1068$ samples.
This dataset is recorded using the same sensor and the same indenters as used during training data collection, i.e., the train indenters (cf. Fig.~\ref{fig:indenters}).
The value ranges of the ground truth force distribution in the test dataset for the Total Force (TF) estimation task are shown in Fig.~\ref{fig:enter-label}.
Given the nature of the indentation experiments, the GelSight Mini sensor therefore primarily encounters normal forces, while the range of the encountered shear forces is smaller.

An example of the model prediction is shown in Fig.~\ref{fig:pred_label_grid}, resulting from the contact with a sloping cuboid indenter.
The U-net demonstrates a noteworthy capability to predict shear and normal force distributions which closely align with the ground truth data (cf. left vs. right column in Fig.~\ref{fig:pred_label_grid}).
This visual comparison highlights the U-net's effectiveness in capturing the fundamental structures of the underlying force distribution patterns.
Additional examples that underscore this finding are provided on the accompanying website.

Quantitative results are also provided in Table~\ref{tab:evaluation_unet}. The U-net is capable of accurately predicting the total shear and normal forces, with mean MAE values below $\unit[1]{N}$. It has been shown that the U-net performs better when dealing with shear forces in the $x$- and $y$-directions than with normal forces in the $z$-direction. This difference in performance may be attributed to two factors. First, the range of force values encountered by the U-net for shear forces are generally smaller than for normal forces. Second, the markers in the GelSight mini sensor's gel inherently encompass a richer set of features for extracting and representing shear behavior. Shear forces cause a greater displacement of the markers compared to normal forces, and might therefore trigger a clearer signal.

\begin{table}[t]
\vspace{1em}
\centering
\caption{
Model Ablation on the Total Force Estimation Task. Colors indicate the three best ranked methods - \firstrank{blue} indicates the best, \secondrank{orange} the second best and \thirdrank{violet} the third best.
}
\label{tab:comparison_unet}
\scalebox{0.9}{
\begin{tabular}{l|ccc}
\multicolumn{1}{c|}{\multirow{2}{*}{Method}} & \multicolumn{3}{c}{$\text{MAE}_{\text{TF}}$ {[}N{]}} \\
 & $f_x$ & $f_y$ & $f_z$ \\ \hline
$\text{ResNet-34}^{\text{\tiny $1\stimes3$}}$ & \scriptsize $\firstrank{{0.077 \pm 0.124}}$ & \scriptsize $\firstrank{0.083 \pm 0.115}$ & \scriptsize $1.513 \pm 1.186$ \\
$\text{U-net}^{\text{\tiny $12\stimes16\stimes3$}}$ & \scriptsize \secondrank{$0.206 \pm 0.307$} & \scriptsize \secondrank{$0.163 \pm 0.242$} & \scriptsize \firstrank{$0.761 \pm 0.941$} \\
$3\stimes\text{U-net}^{\text{\tiny $24\stimes32\stimes1$}}$ & \scriptsize $0.426 \pm 0.521$ & \scriptsize $0.401 \pm 0.572$ & \scriptsize $\secondrank{0.783 \pm 1.022}$ \\
$\textbf{\text{U-net}}^{\text{\tiny $24\stimes32\stimes3$}}$ \textbf{(ours)} & \scriptsize $\thirdrank{0.293 \pm 0.417}$ & \scriptsize $\thirdrank{0.230 \pm 0.341}$ & \scriptsize $\thirdrank{0.799 \pm 1.014}$ \\
$\text{U-net}^{\text{\tiny $48\stimes64\stimes3$}}$ & \scriptsize $0.338 \pm 0.484$ & \scriptsize $0.272 \pm 0.392$ & \scriptsize $1.037 \pm 1.287$ \\ \hline
\end{tabular}
}
\end{table}

\paragraph{Impact of Label Resolution on Prediction Accuracy}
This section compares our main U-net model with variants of the U-net architecture and against a ResNet to investigate whether changing the output resolution and network structure affects the prediction quality.
We consider the same testing dataset as in the previous experiment.

The results on the Total Force (TF) estimation task in Table~\ref{tab:comparison_unet} show that the ResNet, which solely regresses to the total force, outperforms all considered U-net architectures in predicting the shear forces but is significantly less accurate in predicting the normal forces.
Importantly, the ResNet only outputs a single 3D vector of force for the whole sensor, and therefore significantly lacks spatial resolution.
The top performing model according to the MAE on the total force is the U-net with output dimension of $12 \times 16 \times 3$.
Within the variance of the results, it is however very close to our main architecture with output $24 \times 32 \times 3$.
Since we aim for a balance between resolution and achieving small errors on the test dataset, we therefore selected the model with the output resolution of $24 \times 32 \times 3$.

We also compare against training three separate U-net models, where each model is designed to solely predict a single force component, i.e., each individual model outputs a force distribution of shape $24 \times 32 \times 1$.
Although these models collectively possess three times the number of parameters, their predictive accuracy is inferior to that of the main U-net w.r.t. the shear force estimates and only minorly improved w.r.t. normal force. This suggests that predicting all three force directions simultaneously can better capture correlations, especially between the shear forces, yielding more accurate predictions. Moreover, using one joint model also drastically improves the computational efficiency.

Lastly, when further increasing the output resolution to $48 \times 64 \times 3$, we still observe good performance; however, a slight decline compared to our main U-net.

\paragraph{Generalization to Novel Test Indenters}

\begin{table}[t]
\vspace{1em}
\centering
\caption{U-net Mean Absolute Error (MAE) on the Test Set containing new, previously unseen Indenters.}
\label{tab:evaluation_unet_unknown_indenters}
\begin{tabular}{lcc}
\multicolumn{1}{c|}{} & $\text{MAE}_{\text{GUF}}$ {[}N{]} & $\text{MAE}_{\text{TF}}$ {[}N{]} \\ \hline
\multicolumn{1}{l|}{$f_x$} & $0.0014 \pm 0.0014$ & $0.3010 \pm 0.3777$ \\
\multicolumn{1}{l|}{$f_y$} & $0.0013 \pm 0.0014$ & $0.2193 \pm 0.2837$ \\
\multicolumn{1}{l|}{$f_z$} & $0.0042 \pm 0.0032$ & $1.1862 \pm 1.2998$ \\ \hline
\end{tabular}
\end{table}

While the previous experiment evaluated our trained representation on the same indenters that were also used for training, this section now evaluates FEATS on $5$ new, previously unseen test indenters which were not included in the training data (cf. Fig.~\ref{fig:indenters}).
We collected $2789$ samples with these new indenters, covering the same force range as for the other testing dataset (cf. Fig.~\ref{fig:enter-label}).
The results in Tab.~\ref{tab:evaluation_unet_unknown_indenters} reveal that our trained network is indeed capable of generalizing to the previously unseen indenters.
Considering the per-Grid Unit Forces (GUF), the errors only slightly increase by $\unit[0.0004]{N}$, $\unit[0.0003]{N}$, for the x- and y-direction, while the total force errors remain comparable to the previous experiment.
In line with the higher range of the normal forces in this dataset, we also observe a slightly increased error of $\unit[0.0012]{N}$ compared to the evaluation on the known indenters.
Since we only observe a small increase in the errors compared to the known indenters, we conclude that the representation generalizes well to these new contact configurations.

\begin{figure}[t]
    \centering
    \rotatebox[origin=l]{90}{\makebox[0.65in]{\:}}%
    \hskip 0.1em
    \subfloat{\includegraphics[height=4.4em]{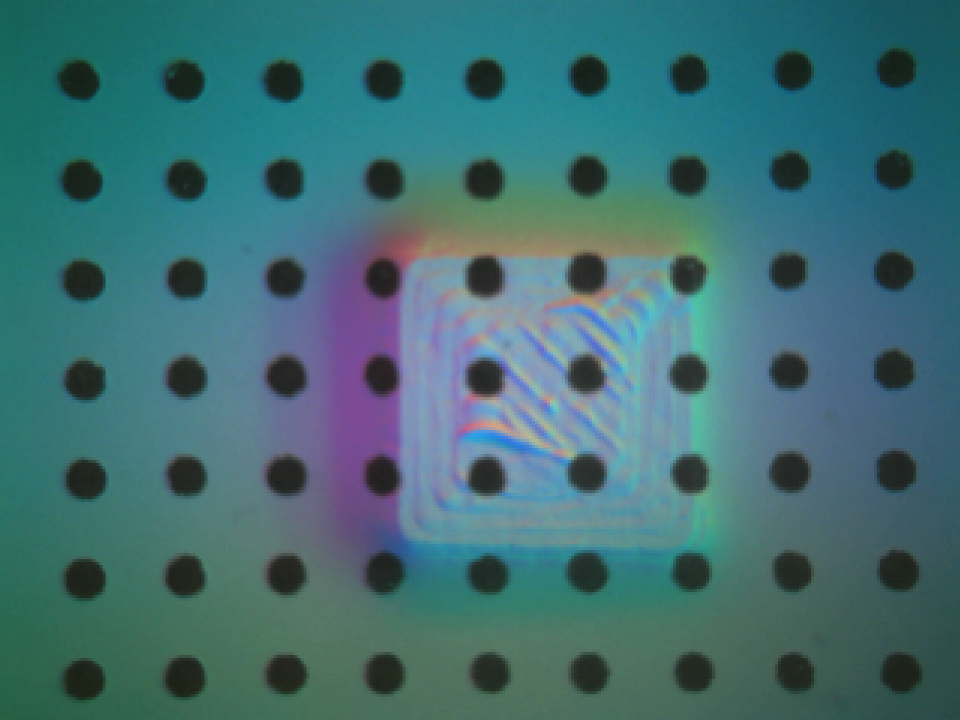}} \hskip 1.55em
    \subfloat{\includegraphics[height=4.4em]{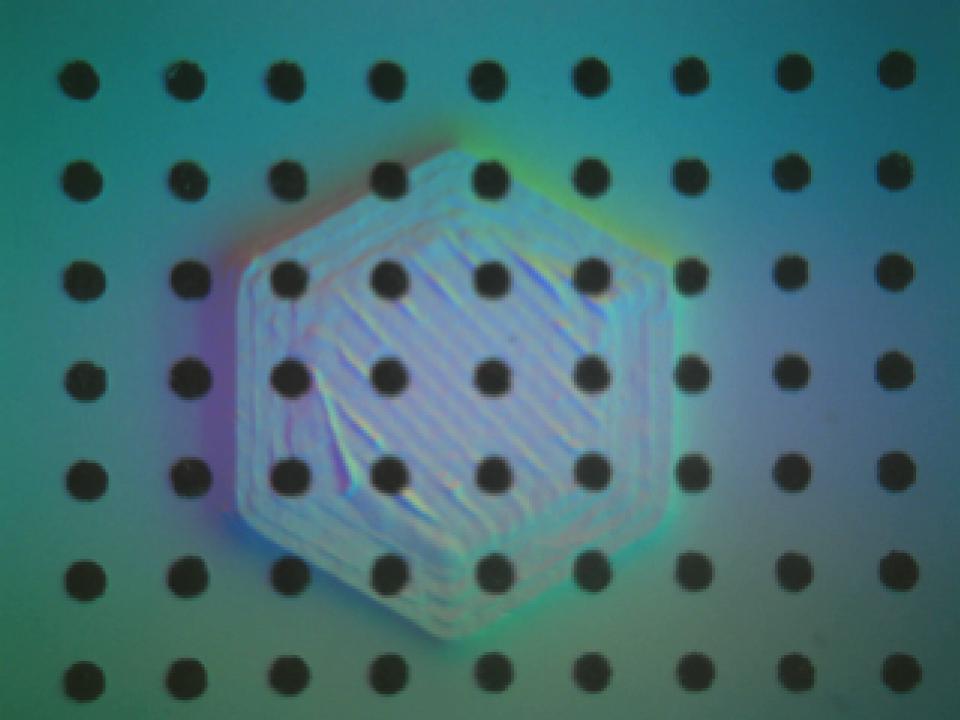}} \hskip 1.55em
    \subfloat{\includegraphics[height=4.4em]{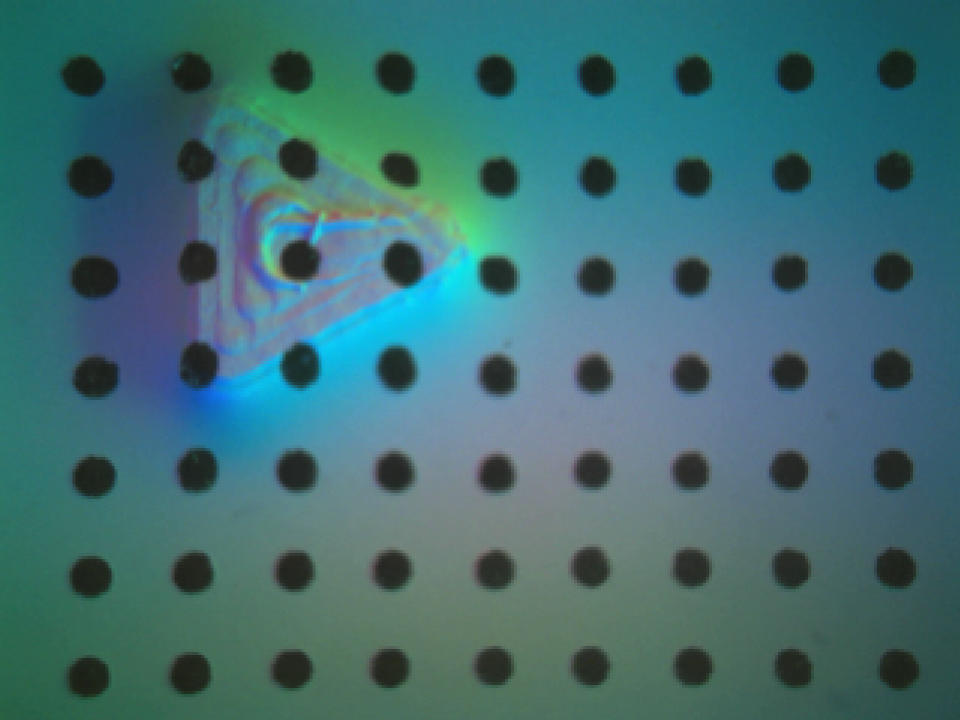}} \vspace{-0.5em}
    
    \rotatebox[origin=l]{90}{\makebox[0.65in]{\scriptsize $x$ component}}%
    \hskip 0.2em
    \subfloat{\includegraphics[height=4.8em]{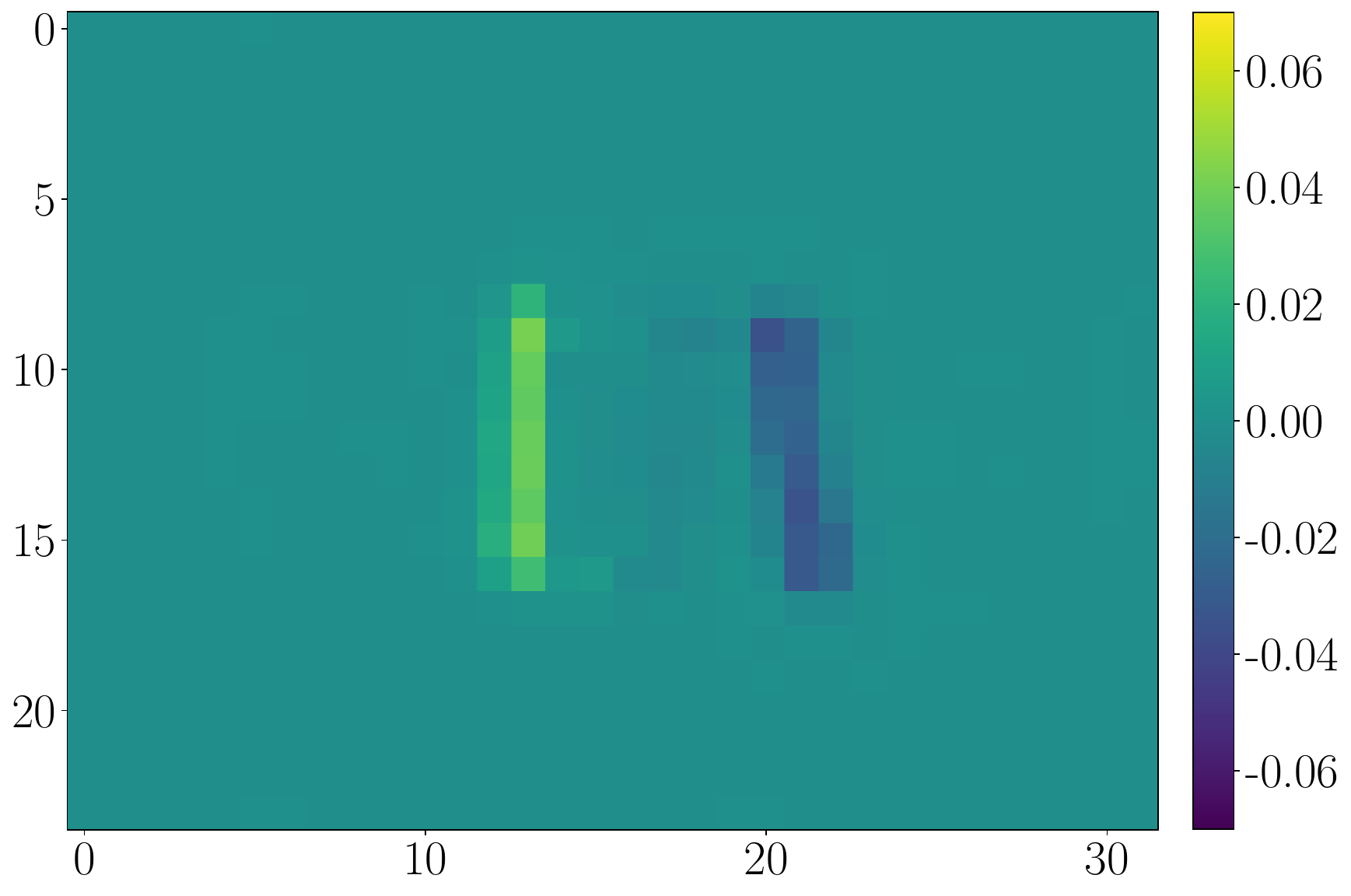}} \hskip 0.2em
    \subfloat{\includegraphics[height=4.8em]{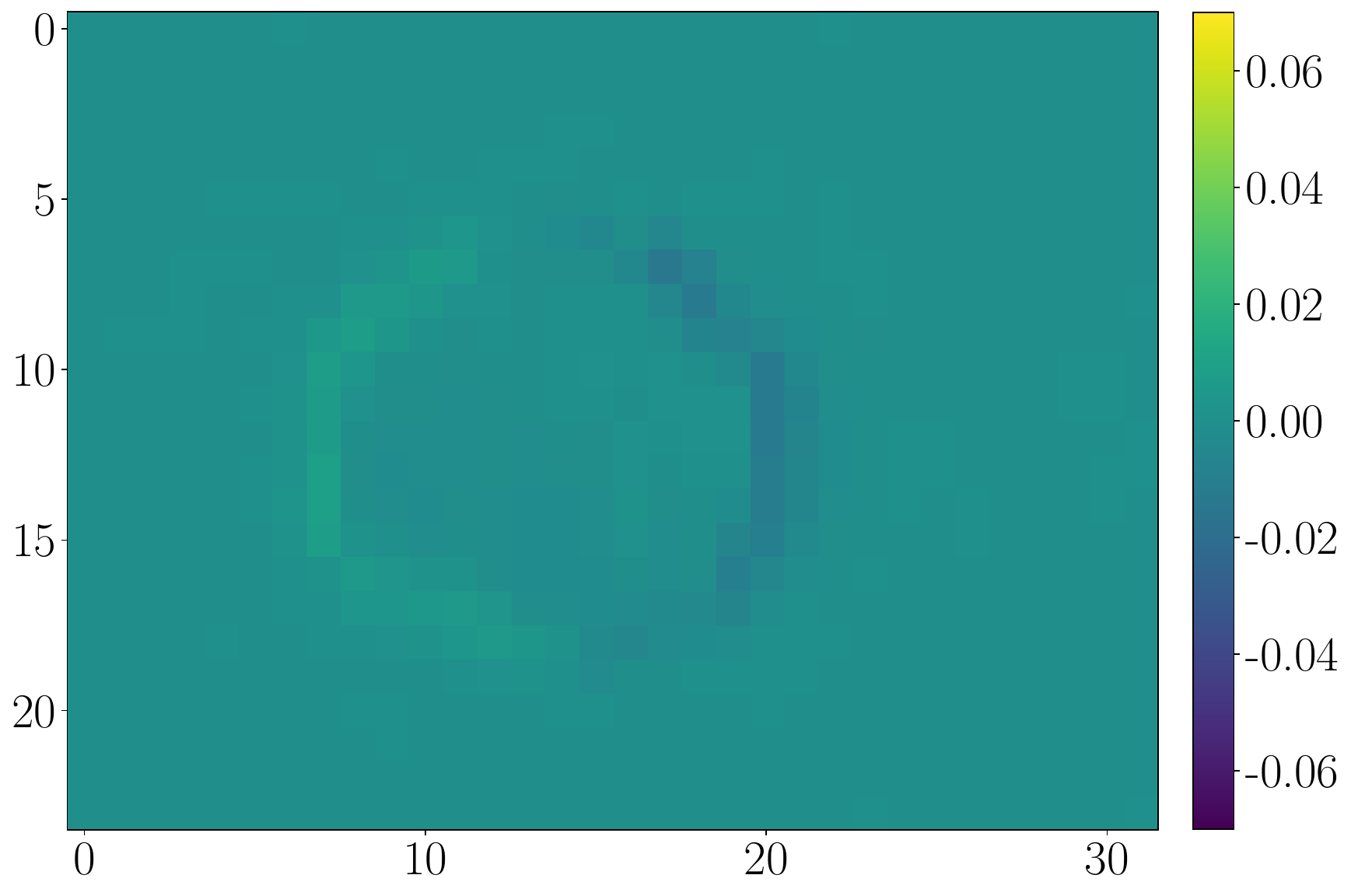}} \hskip 0.2em
    \subfloat{\includegraphics[height=4.8em]{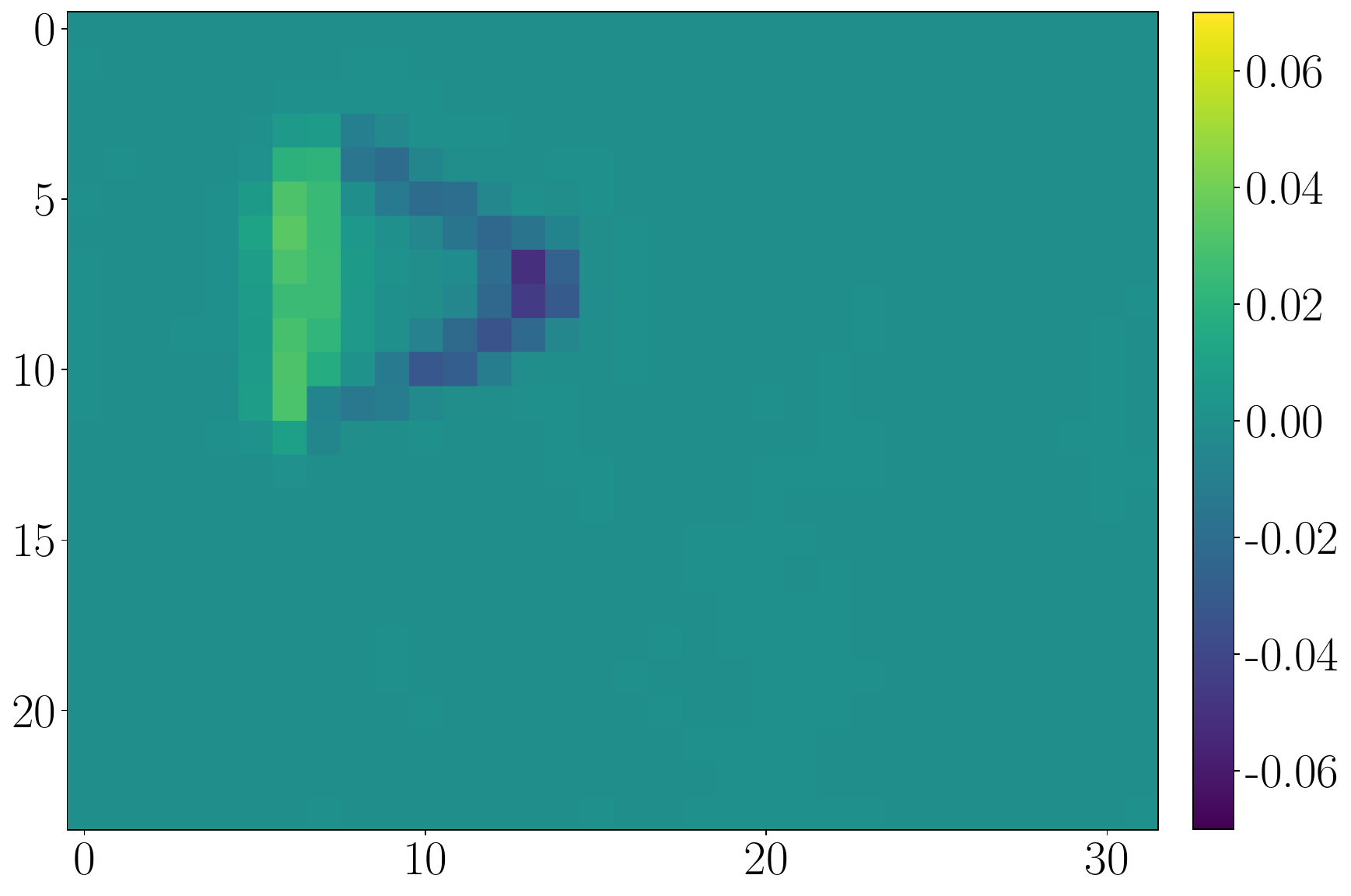}} \vspace{-0.5em}

    \rotatebox[origin=l]{90}{\makebox[0.65in]{\scriptsize $y$ component}}%
    \hskip 0.2em
    \subfloat{\includegraphics[height=4.8em]{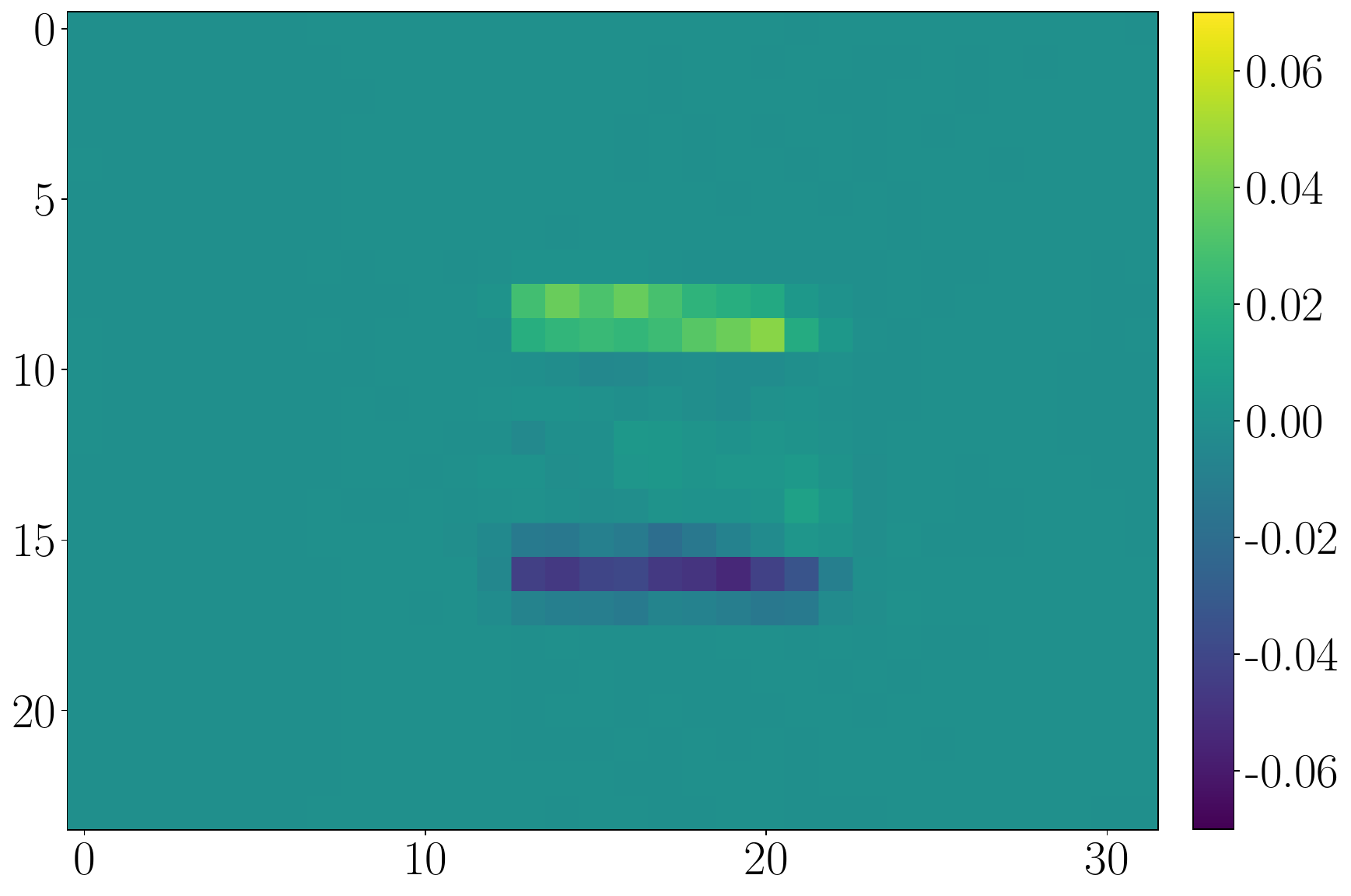}} \hskip 0.2em
    \subfloat{\includegraphics[height=4.8em]{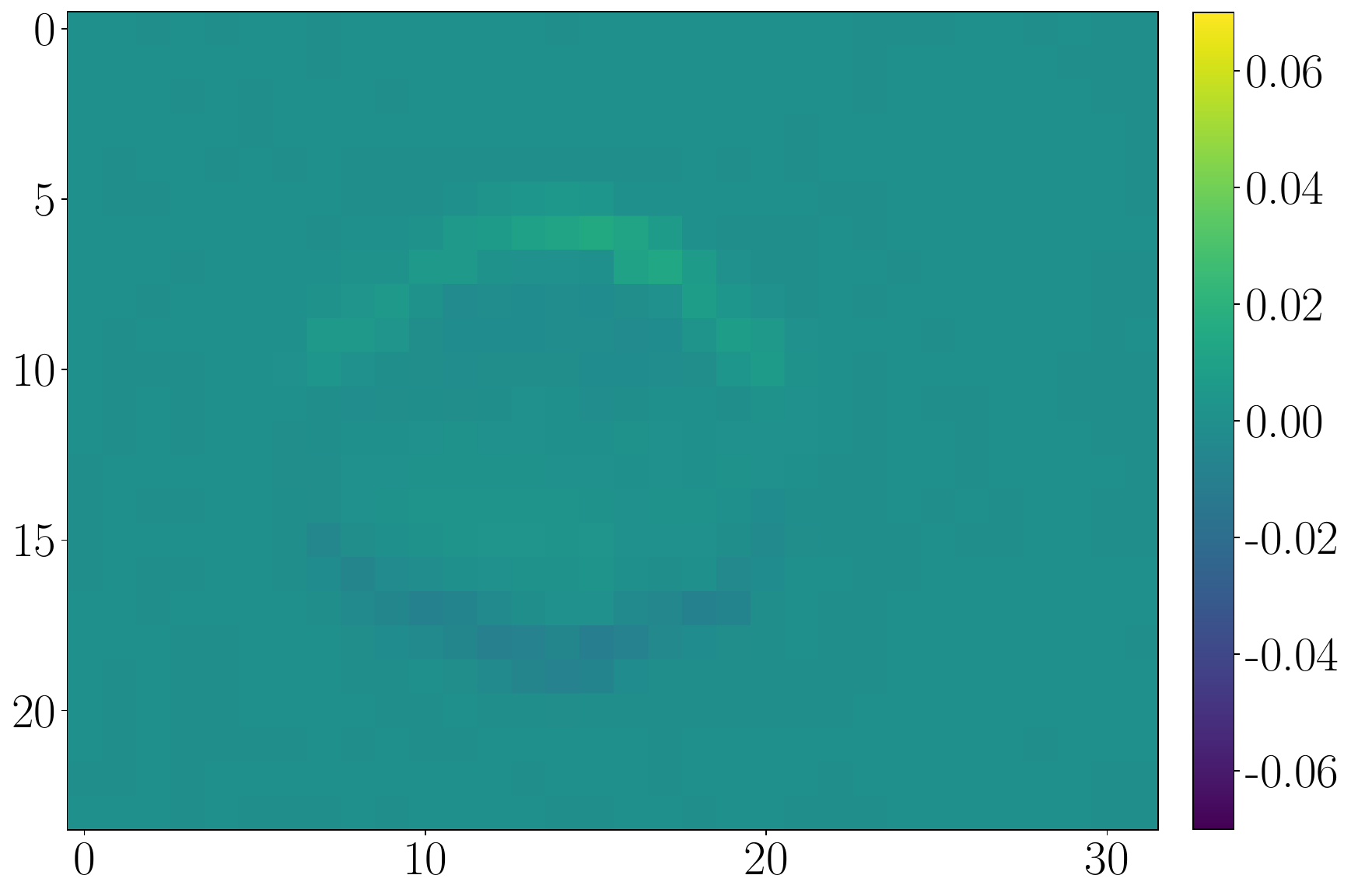}} \hskip 0.2em
    \subfloat{\includegraphics[height=4.8em]{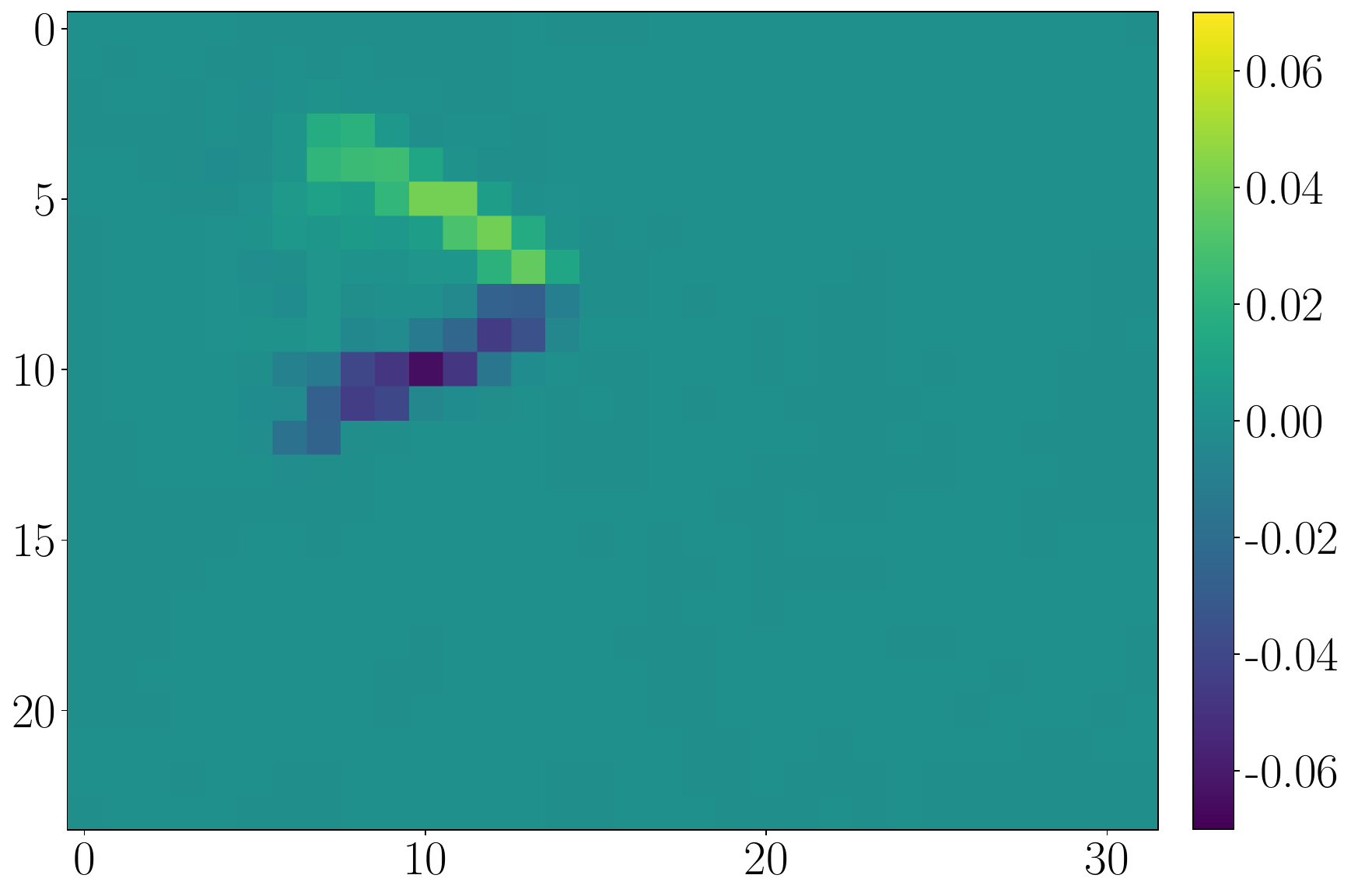}} \vspace{-0.5em}

    \rotatebox[origin=l]{90}{\makebox[0.65in]{\scriptsize $z$ component}}%
    \hskip 0.2em
    \subfloat{\includegraphics[height=4.9em]{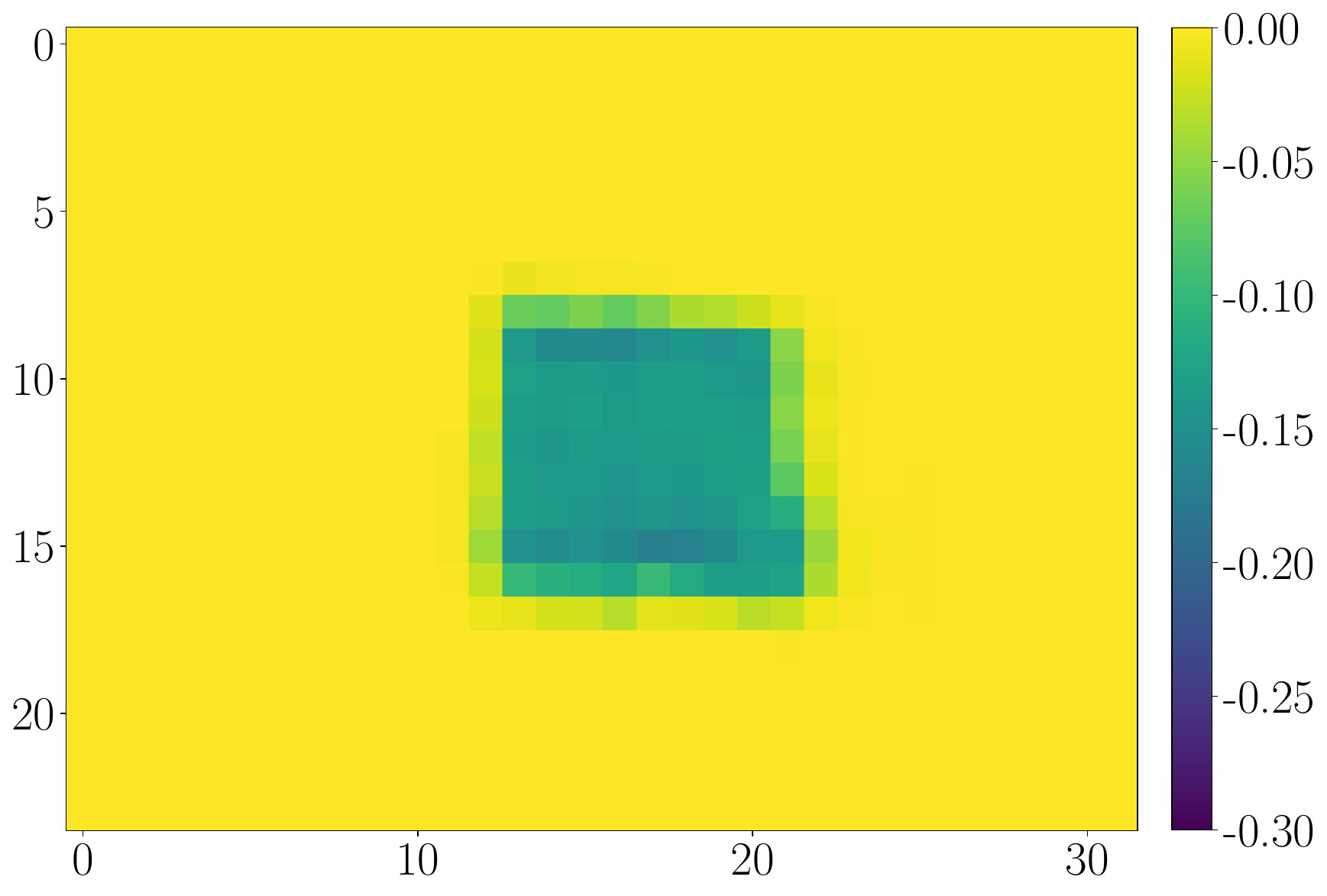}} \hskip 0.2em
    \subfloat{\includegraphics[height=4.9em]{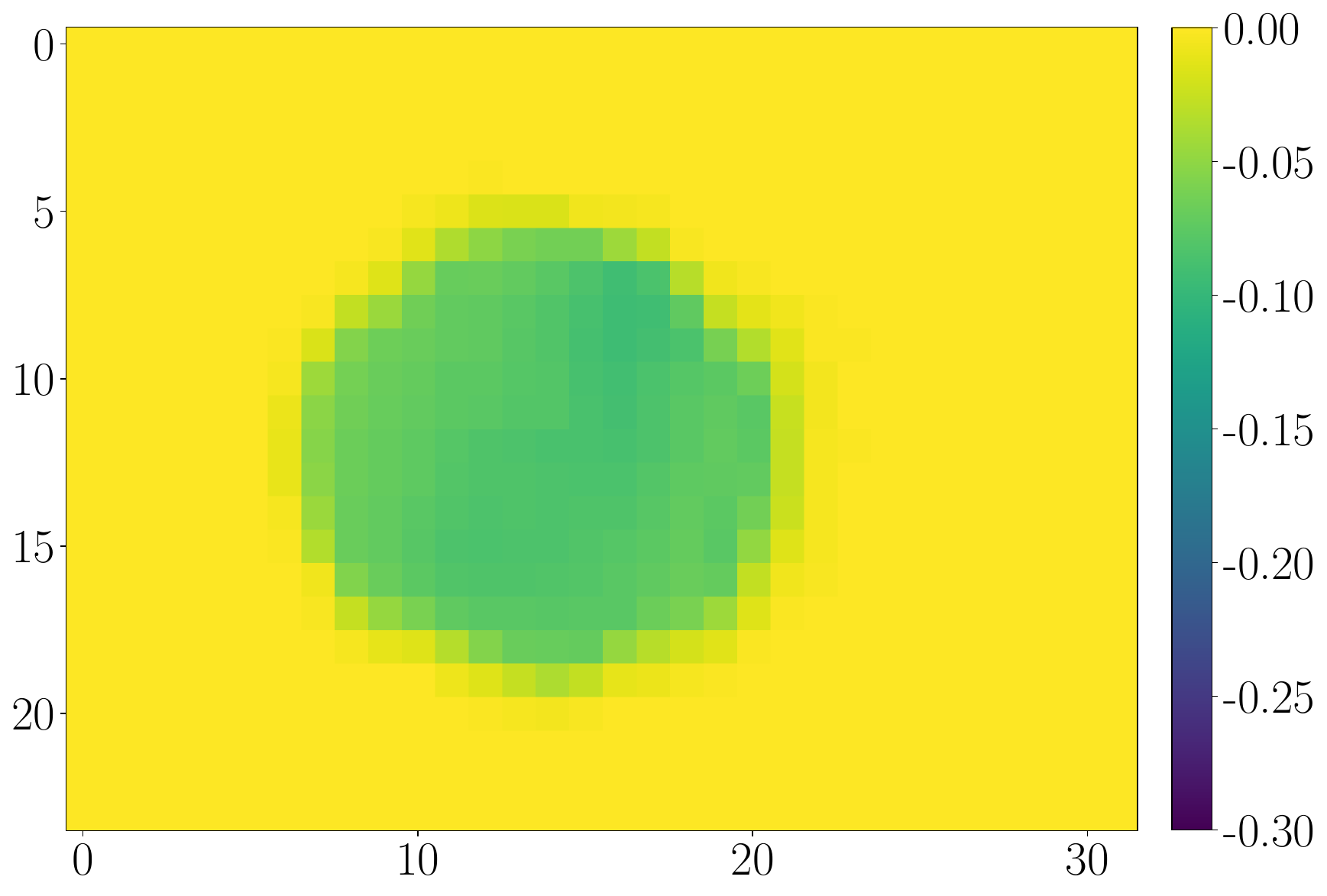}} \hskip 0.2em
    \subfloat{\includegraphics[height=4.9em]{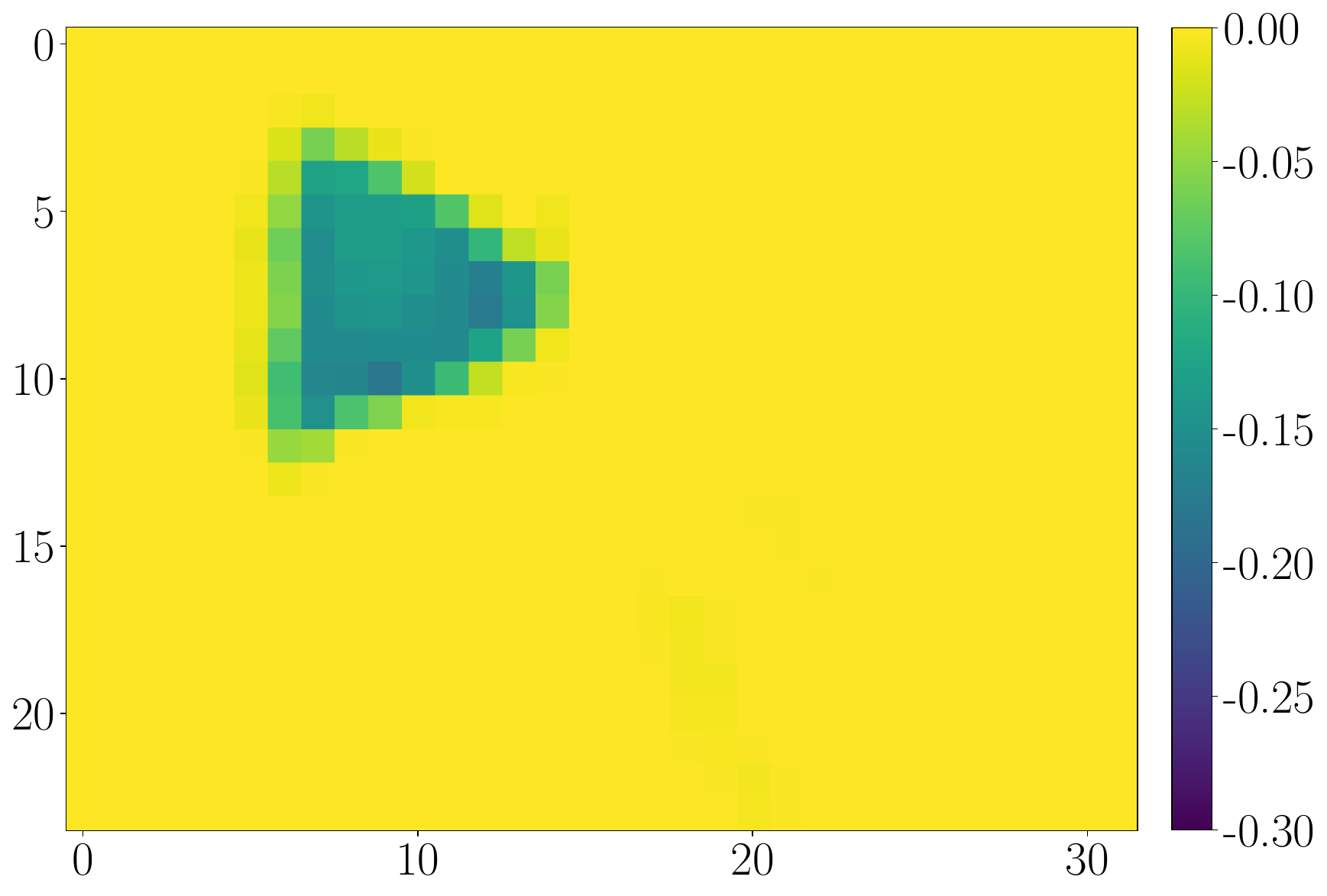}}
    
    \caption{Generalization to a different sensor.
    We evaluate our FEATS U-net on the previously unseen test indenters using a different GelSight Mini sensor to validate the model generalization capability.
    The first row displays the raw images captured by the sensor, and the subsequent rows show the components of the force distribution, in Newtons.
    We observe that the general outline of the force distribution is captured correctly despite the new sensor and the previously unseen indenters.
    }
    \label{fig:different_sensor_newGel}
\end{figure}

\paragraph{Generalization to Different Sensors}
To explore the generalization capability of our FEATS method, we finally test it on images captured with a different GelSight Mini sensor of the same type and a new gel. We again consider the previously unseen test indenters. Initially, we obtain mean total force errors of $\unit[0.24]{N}$, $\unit[0.27]{N}$, \& $\unit[2.99]{N}$ for the $x$-, \mbox{$y$-,} \& $z$-directions, respectively. To counteract the increased errors compared to the previous evaluation, we collected $4400$ additional samples from four GelSight Minis, systematically swapping their gels, and integrated these samples into our training dataset. The additional data is intended to capture the variations among different GelSight Mini sensors.
This significantly enhanced generalization, achieving reduced errors of $\unit[0.23]{N}$, $\unit[0.25]{N}$, \& $\unit[1.79]{N}$ for a previously unseen sensor-gel combination (see Fig.~\ref{fig:different_sensor_newGel}).
The larger increase in error for the normal force ($z$-direction) might be attributed to factors like camera alignment and sensor-specific image variations. A demonstration video showing a dynamic interaction is available on our project website.

\subsection{Inference Speed}
\label{sec:inference_speed}
We finally report our method's inference time on a computer equipped with a \textit{NVIDIA Quadro RTX 5000} and an \textit{Intel Core i7-10875H 8-Core} CPU.
The time for transferring data between the CPU and GPU, as well as asynchronous execution and warm-up of the GPU is taken into consideration. On average, the U-net achieves an inference time of $\unit[4.1746 \pm 1.339]{ms}$ over $300$ runs, which is sufficient for obtaining the force distributions in real-time since the GelSight Mini sensor runs at $\unit[25]{Hz}$.
This is significantly faster compared to running the FEA simulations, which may take $10 - 120$ minutes depending on the contact geometry.

\section{Conclusion}
We introduced FEATS---a machine learning approach for estimating force distributions using the GelSight Mini tactile sensor.
By training a U-net model on \ac{fea}-derived data, we achieved accurate predictions of both shear and normal forces from raw sensor images, with Mean Absolute Error (MAE) under~$\unit[0.8]{N}$ on average in the range $\unit[0 - 40]{N}$ for the total normal force and under~$\unit[0.3]{N}$ on average for the total shear force in each direction with a range of $\unit[-5]{N}$ to $\unit[5]{N}$.
The model showed potential for real-time applications and generalization to previously unseen indenters and GelSight Mini sensors, which could be further improved through domain adaptation~\cite{chen2024deep}.
Our FEATS method offers an efficient online approach that amortizes the cost of running inverse \ac{fea} into a forward pass of a neural network, resulting in 
physically-grounded, interpretable representations for optical tactile sensors.
Future work will focus on enhancing the model’s robustness to different tactile scenarios and work towards demonstrating the representation's effectiveness in robotic manipulation tasks.

\section*{Acknowledgement}
We thank $\text{LAB}^3$, Kai Ruf \& Uwe Faltermeier for their great support and access to the \ac{cnc} milling machine,
and we acknowledge the computing time provided on the Lichtenberg cluster at TU Darmstadt.

\bibliographystyle{IEEEtran}
\bibliography{bibliography}

\end{document}

%% file: includes.tex
\usepackage{blindtext}
\usepackage{float}
\usepackage{amsmath}

\usepackage{mathtools}
\usepackage{array, makecell}
\usepackage{multirow}
\usepackage{csquotes}
\usepackage{graphicx}
\usepackage{caption}
\usepackage{subfig}

\usepackage{units}

\usepackage[dvipsnames]{xcolor}
\usepackage{color, soul}
\usepackage{acronym}
\definecolor{bananamania}{rgb}{0.98, 0.91, 0.71}
\definecolor{buff}{rgb}{0.94, 0.86, 0.51}
\definecolor{celadon}{rgb}{0.67, 0.88, 0.69}
\definecolor{chartreuse(traditional)}{rgb}{0.87, 1.0, 0.0}

\sethlcolor{chartreuse(traditional)}

\usepackage[hidelinks, colorlinks=true, citecolor=blue]{hyperref}

\usepackage{multicol}

\newcommand{\stimes}{{\times}}

%% file: acronyms.tex
\newacro{fem}[FEM]{Finite Element Method}
\newacro{fea}[FEA]{Finite Element Analysis}
\newacro{mdm}[MDM]{Marker Displacement Method}
\newacro{ml}[ML]{Machine Learning}
\newacro{cnn}[CNN]{Convolutional Neural Network}
\newacro{lstm}[LSTM]{Long Short-Term Memory}
\newacro{svm}[SVM]{Support Vector Machine}
\newacro{dnn}[DNN]{Deep Neural Network}
\newacro{cnc}[CNC]{Computer Numerical Control}
\newacro{mae}[MAE]{Mean Absolute Error}
\newacro{mse}[MSE]{Mean Squared Error}